\documentclass[runningheads]{llncs}

\usepackage{eccv}

\usepackage{eccvabbrv}

\usepackage{graphicx}
\usepackage{booktabs}
\usepackage{algorithm}
\usepackage{algpseudocode}
\usepackage{appendix}
\usepackage{xcolor}

\newcommand{\bluetext}[1]{\textcolor{blue}{#1}}
\newcommand{\greentext}[1]{\textcolor{teal}{#1}}
\newtheorem{assumption}{Assumption}
\usepackage{multirow}
\usepackage[accsupp]{axessibility}  

\usepackage{hyperref}

\usepackage{orcidlink}

\begin{document}

\title{Timestep-Aware Correction for Quantized Diffusion Models} 


\author{Yuzhe Yao\thanks{This work was completed during an internship at SGIT AI Lab, State
Grid Corporation of China.}\inst{1} \and
Feng Tian\inst{1}\and
Jun Chen \thanks{Corresponding Author} \inst{2,3} \and
Haonan Lin\inst{1}\and
Guang Dai\inst{3}\and
Yong Liu \inst{4,3}\and
Jingdong Wang \inst{5}}

\authorrunning{Yao et al.}

\institute{Xi'an Jiaotong University \and
Zhejiang Normal University \and
SGIT AI Lab, State Grid Corporration of China \and
Zhejiang University \and
Baidu Inc\\
\email{\{yuzheyao.22,junc.change,guang.gdai\}@gmail.com, fengtian@mail.xjtu.edu.cn, linhaonan@stu.xjtu.edu.cn, yongliu@iipc.zju.edu.cn, wangjingdong@outlook.com}}

\maketitle

\begin{abstract}
Diffusion models have marked a significant breakthrough in the synthesis of semantically coherent images. However, their extensive noise estimation networks and the iterative generation process limit their wider application, particularly on resource-constrained platforms like mobile devices. Existing post-training quantization (PTQ) methods have managed to compress diffusion models to low precision. Nevertheless, due to the iterative nature of diffusion models, quantization errors tend to accumulate throughout the generation process. This accumulation of error becomes particularly problematic in low-precision scenarios, leading to significant distortions in the generated images. We attribute this accumulation issue to two main causes: error propagation and exposure bias. To address these problems, we propose a timestep-aware correction method for quantized diffusion model, which dynamically corrects the quantization error. By leveraging the proposed method in low-precision diffusion models, substantial enhancement of output quality could be achieved with only negligible computation overhead. Extensive experiments underscore our method's effectiveness and generalizability. By employing the proposed correction strategy, we achieve state-of-the-art (SOTA) results on low-precision models.
\end{abstract}

\section{Introduction}
\label{sec:intro}

Diffusion models (DMs)~\cite{ho2020denoising,song2020denoising,sohldickstein2015deep} have emerged as powerful deep generative models for various applications, including image synthesis, Text-to-Image generation, video generation, and medical image reconstruction~\cite{yang2023diffusion,rombach2022high,zhang2023adding,xie2022measurement,saharia2022palette,saharia2022photorealistic,ruiz2023dreambooth,gu2022vector,dhariwal2021diffusion}. 
Despite their ability to generate images with high fidelity and diversity, diffusion models are impeded by a time-consuming and computationally intensive synthesis process. This complexity arises primarily from two factors. First, diffusion models employ a complex deep neural network, such as U-Net~\cite{ronneberger2015u}, for noise estimation. Second, to maintain the quality of the synthesized images, diffusion models necessitate an iterative process that progressively denoises the input image. This procedure can require up to 1,000 iterations, substantially contributing to the overall computational burden.

\begin{figure}[t]
  \centering
    \includegraphics[width=\textwidth]{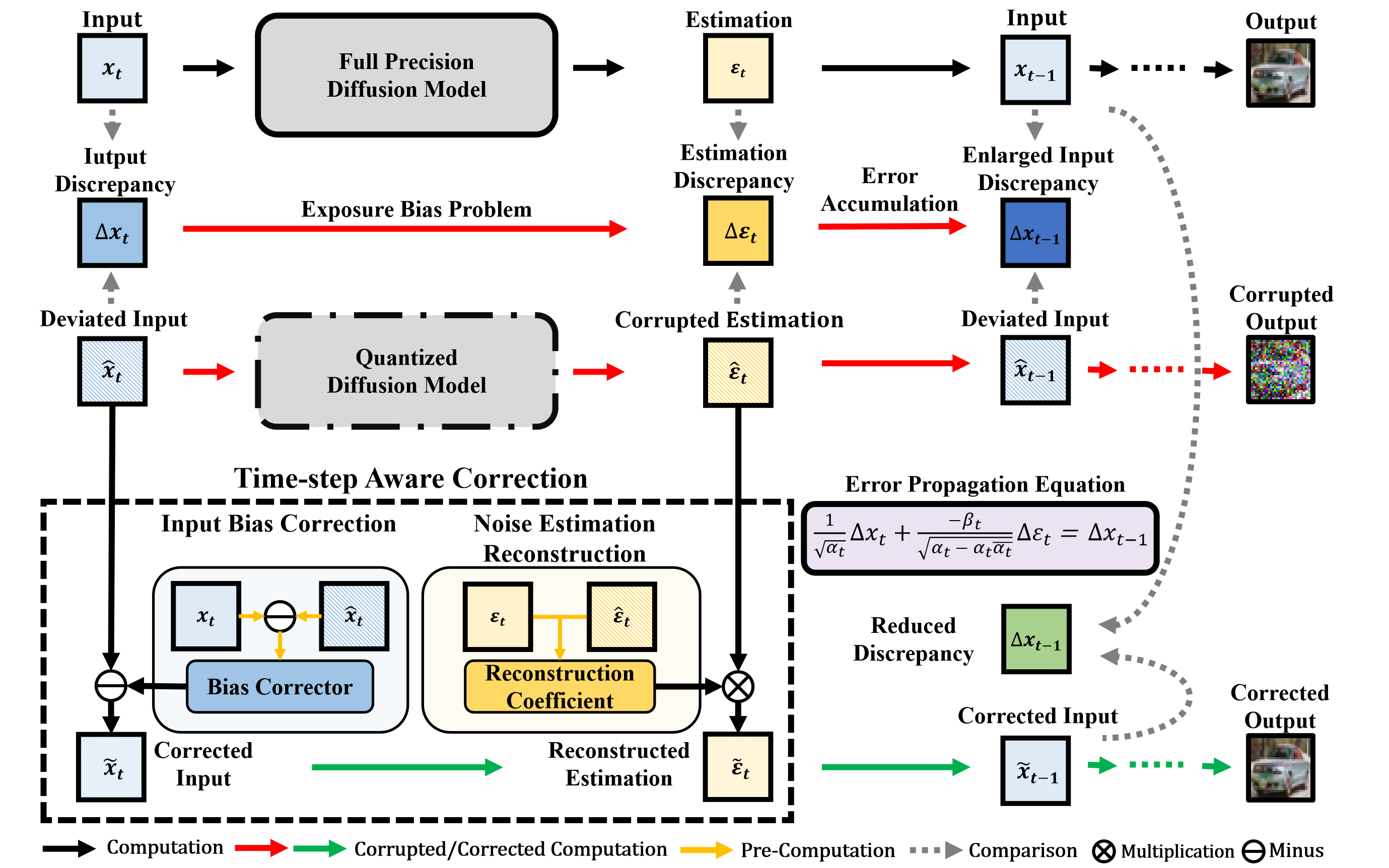}
    \caption{(\textbf{Upper}) Illustration of error accumulation in diffusion models. Inherent to the design of DMs, discrepancy in the input not only propagates to the next timestep but also leads to significant discrepancy in noise estimation, due to exposure bias. This cascading effect amplifies errors in subsequent stages, cumulatively impairing the quality of the final output. (\textbf{Lower-left}) To address this challenge, our method focuses on reducing the accumulated error $\Delta\mathbf{x}_{0}$ through two key strategies: \textit{1)} minimizing the discrepancy $\Delta\mathbf{x}_{t-1}$ at each timestep $t < T$, and \textit{2)} decomposing $\Delta\mathbf{x}_{t-1}$ into two distinct components—the input discrepancy $\Delta\mathbf{x}_t$ and the noise estimation discrepancy $\Delta \boldsymbol{\epsilon}_t$—and rectifying them separately to enhance error correction.}
  \label{fig:eccv_ppl}
  \vspace{-0.6cm}
\end{figure}

In response to the significant computational demands of the diffusion model's noise estimation process, researchers have been pursuing various methods regarding the two factors that slow down the inference. One approach focuses on the refinement of the sampling trajectory, which can effectively shorten the noise estimation phase, as explored in several studies~\cite{song2020denoising,nichol2021improved,lu2022dpm,lu2023dpmsolver,bao2022analyticdpm}. Another approach aims to reduce the latency of the noise estimation network in diffusion models by employing model quantization~\cite{polino2018model}. While Quantization-Aware Training (QAT)~\cite{jacob2018quantization,han2015deep,so2023temporal} requires retraining the neural network with simulated quantization and hyper-parameter search, it is more computation-intensive and requires more engineering effort for deployment~\cite{nagel2021white,gholami2022survey}. In contrast, Post-Training Quantization (PTQ) emerges as a preferred choice for lightweight diffusion models due to its straightforward approach~\cite{shang2023post,li2023q,he2023ptqd}. However, even though PTQ for models like MobileNet\cite{howard2017mobilenets} and ResNet\cite{he2016deep} is a well-studied technique~\cite{nagel2021white,liu2021post,bondarenko2021understanding,nagel2019data,banner2019post}, its application in PTQ in diffusion models warrants further investigation, particularly due to the dynamic nature of DMs.

Recent advancements in PTQ for DMs have enabled the quantization of model weights to 4-bit precision~\cite{huang2023tfmq,li2023q,shang2023post,he2023ptqd}. However, these methods often face challenges in preserving image quality at lower bit-widths (e.g., 3-bit and 2-bit weight parameters), primarily due to error accumulation in DMs~\cite{li2023q,he2023ptqd,li2023qdm}.

We attribute error accumulation in diffusion models to two main factors: error propagation and exposure bias. Mathematical analysis by recent studies~\cite{li2024on} reveals that errors in the inputs of DMs will propagate to subsequent denoising steps. This is particularly problematic in low-precision DMs, where each inference step inherently introduces additional errors, leading to significant cumulative effects. Moreover, the input discrepancy causes exposure bias problem, exacerbating the issue by causing a significant discrepancy in the noise estimation from the ground truth~\cite{schmidt-2019-generalization,ning2024elucidating,ning2023input,li2024alleviating}, resulting in compounded errors in the model's output. The error accumulation challenge is depicted in the upper section of \cref{fig:eccv_ppl}.

To counteract error accumulation, we introduce a timestep-aware correction strategy. This approach, integrating Noise Estimation Reconstruction (NER) and Input Bias Correction (IBC), dynamically mitigates quantization-induced errors at every generation step, significantly improving the fidelity of images produced by low-precision DMs. Importantly, our method does not require additional training or tuning, allowing for seamless integration into existing PTQ frameworks. The operational details of our method are illustrated in the lower-left segment of \cref{fig:eccv_ppl}.

Our study's core contributions are as follows:

\begin{enumerate}
    \item We identify error accumulation as a key challenge in PTQ for diffusion models, attributing it primarily to error propagation and exposure bias. In response, we present Timestep-Aware Correction for Quantized Diffusion Models (TAC-Diffusion), a novel approach for dynamically mitigating errors across the generative process.
    \item We decompose the error in the model's input into two main components: corrupted noise estimation and deviated input from previous timesteps. We propose Noise Estimation Reconstruction (NER) and Input Bias Correction (IBC) to correct each aspect.
    \item Extensive experiments across various diffusion models and samplers for both conditional and unconditional image generation tasks demonstrate our method's effectiveness. Notably, TAC-Diffusion significantly outperforms the state-of-the-art Q-Diffusion method~\cite{li2023q}, achieving a FID improvement of 7.76 on CIFAR-10 with 3-bit weight and 8-bit activation quantization.
\end{enumerate}

\section{Related work}
\label{sec:related_work}

\subsection{Diffusion Models.} 
Diffusion models are latent variable models designed to generate images that match the training set's distribution $q\left(\boldsymbol{x}_0\right)$ from random Gaussian noise~\cite{ho2020denoising,sohldickstein2015deep,song2020denoising,rombach2022high}. The generating process can be modeled as:
\begin{equation}
    p_\theta\left(\mathbf{x}_0\right):=\int p_\theta\left(\mathbf{x}_{0: T}\right) d \mathbf{x}_{1: T},
\end{equation}
where $\mathbf{x}_{0:T}$ represents a sequence of latent variables from the initial state $\mathbf{x}_0$ to the final state $\mathbf{x}_T$, $\theta$ denotes the parameters, $p_\theta\left(\mathbf{x}_0\right)$ denotes the model distribution and  $p_\theta\left(\mathbf{x}_{0: T}\right)$ denotes the reverse process~\cite{song2020denoising}.

The training of diffusion models involves two key processes: the forward process and the reverse process. In the forward process, Gaussian noise, scheduled by the parameter $\beta_t$, is incrementally added to the latent variable $\mathbf{x}_{t-1}$ at each timestep. This process can be mathematically described as follows:
\begin{equation}
    q\left(\mathbf{x}_{t} \mid \mathbf{x}_{t-1}\right) = \mathcal{N}\left(\mathbf{x}_{t} ; \sqrt{1-\beta_{t}} \mathbf{x}_{t-1}, \beta_{t} \mathbf{I}\right).
    \label{eq:3}
\end{equation}

With an appropriately set noise schedule $\beta_t$, the variable $\mathbf{x}_t$ gradually converges to a standard Gaussian distribution as $t$ increases, for all initial values $\mathbf{x}_0 \sim q\left(\mathbf{x}_0\right)$. This convergence is crucial for the reverse process, where the model generates images by sampling noise from Gaussian distribution and iteratively refining the sample through a series of steps to produce the final output image.
In the reverse process, each iteration of DMs can be formulated as:
\begin{equation}
        p_{\theta}\left(\mathbf{x}_{t-1} \mid \mathbf{x}_{t}\right) = \mathcal{N}\left(\mathbf{x}_{t-1} ; \boldsymbol{\mu}_{\theta}\left(\mathbf{x}_{t}, t\right), \boldsymbol{\Sigma}_{\theta}\left(\mathbf{x}_{t}, t\right)\right),
        \label{eq:sampling_t}
\end{equation}
where $p\left(\mathbf{x}_T\right) \sim \mathcal{N}\left(\mathbf{x}_T ; \mathbf{0}, \mathbf{I}\right)$, with
$\boldsymbol{\mu}_{\theta}\left(\mathbf{x}_{t}, t\right)$ denotes the model estimation for noise, and $\boldsymbol{\Sigma}_{\theta}\left(\mathbf{x}_{t}, t\right)$ denotes the variance for sampling which can be fixed to constants~\cite{ho2020denoising}. When the training process ends, \cref{eq:sampling_t} is used for inference. While in DDPM~\cite{ho2020denoising}, the noise estimation process is a Markov process, the inference of diffusion models is inevitably time-consuming and computationally resource-intensive.
\subsection{Post-Training Quantization.} 

Post-Training Quantization (PTQ) is a prominent technique for compressing neural network models to reduce inference latency~\cite{nagel2021white,chen2023data,Bai_2023_ICCV,bondarenko2021understanding,liu2021post}. PTQ converts full-precision (typically FP32) models to a fixed-point format by reducing the bit-width of weights and activations. The method comprises two principal operations: quantization, which utilizes a lower bit width to enhance computation efficiency, and de-quantization, which reverses this process for operations that necessitate floating-point accuracy. The processes are defined as follows:

\begin{equation}
    X_{q}= \operatorname{Clip}\left(\operatorname{Round}\left(\frac{X_{fp}}{s}\right)-z, q_{\min}, q_{\max}\right), \quad X_{deq} = s\left(X_{q} + z\right),
\end{equation}
where \(s\) is the scaling factor determined by the range of full-precision data, \(z\) represents the zero point, and \(q_{max}\) and \(q_{min}\) denote the quantization thresholds.

Calibration, vital for determining the optimal quantization thresholds (\(q_{max}\) and \(q_{min}\)), ensures that the quantized output retains the most significant data~\cite{hubara2020improving}. During this process, PTQ introduces two types of errors: rounding error and clamping error. Rounding error occurs due to the quantization of continuous values to discrete counterparts, while clamping error results from constraining values within the specified \(q_{min}\) and \(q_{max}\) thresholds to prevent overflow or underflow. Efforts to mitigate rounding errors include task-specific rounding strategy~\cite{nagel2020up} and block reconstruction~\cite{li2021brecq}. To address clamping error, Esser\etal~\cite{esser2020learned} carefully selects the quantization threshold during calibration.
\raggedbottom

\subsection{Challenges in PTQ for Diffusion Models}

One primary challenge in PTQ for diffusion models is the variability of activation ranges, where differences across timesteps lead to clamping errors due to the use of a fixed quantization step size \(s\). In response, Shang et al.~\cite{shang2023post} devised a calibration set that specifically incorporates samples near \(t=0\), using a skew-normal distribution. Li et al.~\cite{li2023q} introduced uniform timestep sampling and shortcut-splitting quantization as strategies to alleviate quantization errors in UNet structures. He et al.~\cite{he2023ptqd} recommended varying the bit-widths for activation quantization across timesteps, coupled with variance re-scheduling, to effectively manage quantization errors. Similarly, Ho et al.~\cite{so2023temporal} introduced a novel method that dynamically adjusts the quantization interval based on timestep information. These innovative approaches have successfully preserved the performance of quantized DMs with 4-bit weights and 8-bit activations.

Despite the advancements in PTQ for DMs, another challenge, the error accumulation, remains underexplored. Error accumulation refers to a phenomenon where initial errors propagate and amplify in later timesteps due to exposure bias~\cite{li2024alleviating,ning2023input,ning2024elucidating,li2024on}. In our work, by addressing this key issue during models' inference, we aim to close the gap between quantized diffusion models and their full-precision counterparts.

\section{Problem Assumption and Statement}
\label{sec:problem_statement}
\textbf{Notation} Given a full-precision noise estimation network in diffusion models, $\boldsymbol{\epsilon}_{\theta}$, we denote its quantized version as $\boldsymbol{\hat{\epsilon}}_{\theta}$, their input latent variable at timestep $t$ as $\mathbf{x}_t$ and $\hat{\mathbf{x}}_t$, their noise estimation as $\boldsymbol{\epsilon}_{\theta}\left(\mathbf{x}_{t},t \right)$ and $ \hat{\boldsymbol{\epsilon}}_{\theta}\left(\mathbf{\hat{x}}_{t},t \right)$, respectively. 

In DDPM~\cite{ho2020denoising}, the input of a full-precision noise estimation network at time step $t-1$ can be formulated as:
\begin{equation}
\mathbf{x}_{t-1}=\frac{1}{\sqrt{\alpha_{t}}}\left(\mathbf{x}_{t}-\frac{\beta_{t}}{\sqrt{1-\bar{\alpha}_{t}}} \boldsymbol{\epsilon}_{\theta}\left(\mathbf{x}_{t}, t\right)\right)+\sigma_{t} \mathbf{z},
\label{eq:8}
\end{equation}
where $\alpha_{t}$, $\beta_{t}$, $\Bar{\alpha}_{t}$ and $\sigma_{t}$ are fixed full-precision constants, and $\mathbf{z} \sim \mathcal{N}(\mathbf{0}, \mathbf{I})$. We define the error on the input of a quantized model at timestep $t-1$ as:

\begin{equation}
\begin{split}
 \Delta \mathbf{x}_{t-1} 
 &= \hat{\mathbf{x}}_{t-1} - \mathbf{x}_{t-1} \\
 &= \frac{1}{\sqrt{\alpha_{t}}}\left(\mathbf{\hat{x}}_{t}- \mathbf{x}_{t} \right) - \frac{\beta_{t}}{\sqrt{\alpha_{t}- \alpha_{t}\bar{\alpha}_{t}}} \left(\boldsymbol{\hat{\epsilon}}_{\theta}\left(\mathbf{\hat{x}}_{t}, t\right)- \boldsymbol{\epsilon}_{\theta}\left(\mathbf{x}_{t},t\right)\right) \\
 &=  \frac{1}{\sqrt{\alpha_{t}}}\Delta \mathbf{x}_{t} -  \frac{\beta_{t}}{\sqrt{\alpha_{t}- \alpha_{t}\bar{\alpha}_{t}}}\Delta \mathbf{\boldsymbol{\epsilon}}_{t}.
\end{split}
\label{eq:integrated_noise}
\end{equation}
Based on the above decomposition, we can deduce that the input error at $t-1$ stems from two sources: the input discrepancy and the noise estimation discrepancy at the previous timestep, represented by $\Delta \mathbf{x}_{t}$ and $\Delta \mathbf{\boldsymbol{\epsilon}}_{t}$, respectively. Therefore any input discrepancy will propagate to next timestep. Since the noise estimation network $\epsilon_{\theta}$ is non-linear, and the quantized model $\boldsymbol{\hat{\epsilon}}_{\theta}$ has a different input-output mapping compared to its full-precision version $\boldsymbol{\epsilon}_{\theta}$, it is difficult to predict the input error's influence on the noise estimation. As a consequence, we propose to minimize these two terms separately. We have:

\begin{equation}
\begin{split}
 \|\Delta \mathbf{x}_{t-1}\|
 &= \|\frac{1}{\sqrt{\alpha_{t}}}\Delta \mathbf{x}_{t} -  \frac{\beta_{t}}{\sqrt{\alpha_{t}- \alpha_{t}\bar{\alpha}_{t}}}\Delta \mathbf{\boldsymbol{\epsilon}}_{t}\| \\
 &\leq \frac{1}{\sqrt{\alpha_{t}}}\|\Delta \mathbf{x}_{t}\| + \frac{\beta_{t}}{\sqrt{\alpha_{t}- \alpha_{t}\bar{\alpha}_{t}}} \|\Delta\boldsymbol{\epsilon}_t\|.
\end{split}
\label{eq:relax}
\end{equation}

Above formulation indicates that, if we can effectively rectify $\Delta \mathbf{x}_t$ and $\Delta \mathbf{\boldsymbol{\epsilon}}_t$, we might success in reducing $\Delta \mathbf{x}_{t-1}$. This further leads us to one solution for the error accumulation challenge in quantized DMs: by rectifying the error in the input and the noise estimation for all timestep, the error accumulated at the end $\Delta \mathbf{x}_{0}$ shall be reduced, which indicates that high-quality synthesis images would be obtainable with quantized diffusion models.

\begin{assumption}\label{assumption:assu1} ~\cite{li2024on} Given a quantized diffusion model and its full-precision version, at timestep $t$, reducing their input discrepancy $\Delta\mathbf{x}_t$, and their noise estimation discrepancy, $\Delta \boldsymbol{\epsilon}_{t}$, is to reduce the discrepancy between their final outputs, $\Delta \mathbf{x}_{0}$.
\end{assumption}

Consequently, we propose to reduce the input discrepancy and the noise estimation discrepancy separately for each timestep.

\section{Timestep-Aware Correction}
\label{sec:method}
In this section, we present the timestep-aware correction method, which dynamically reduces the quantization error accumulated on the latent image. At each timestep, the correction consists of two processes: \textit{\textbf{noise estimation reconstruction}} (NER) and \textbf{\textit{input bias correction}} (IBC). With the proposed method, we can effectively correct the quantization error and alleviate the error accumulation during the inference of low-precision DMs. 

\subsection{Noise Estimation Reconstruction}
\label{sec:ner}
To reduce the noise estimation discrepancy between the full-precision model and its quantized version, an intuitive method is to introduce an additional neural network to predict error in the corrupted estimation. However, the complexity of this approach is not consistent with the simplicity of PTQ. Therefore, we aim to find a simple yet effective way to reconstruct the noise estimation. 

Previous work in PTQ~\cite{Bai_2023_ICCV,chen2023data} succeeded in reconstructing the activation map in intermediate layers using damaged activation. Inspired by this approach, we assume that the noise estimation from a full-precision model $\boldsymbol{\epsilon}_t$ can be partially reconstructed from its corrupted version $\boldsymbol{\hat{\epsilon}_t}$, with a linear function.

\begin{assumption}
Given a quantized diffusion model, if its input distortion to the full-precision one $\hat{\mathbf{x}}_t - \mathbf{x_t}$ is small enough, \eg$\mathbf{\hat{x}}_{t} \approx \mathbf{x}_t$, the noise estimation from the full-precision diffusion model $\boldsymbol{\epsilon}_{\theta}\left(\mathbf{x}_{t}, t\right)$ can be partially reconstructed by scaling the noise estimation from the quantized model $\boldsymbol{\hat{\epsilon}}_{\theta}\left(\mathbf{\hat{x}}_{t}, t\right)$ with a channel-wise scaling factor.
\end{assumption}

We then introduce the timestep-dependant reconstruction coefficient $\mathbf{K} \in \mathbb{R}^{T \times C}$, where $T$ denotes the number of denoising steps, and $C$ denotes the number of channels of estimated noise. Then the reconstructed noise in timestep $t$ can be represented as follows:
\begin{equation}
    \boldsymbol{\tilde{\epsilon}}_{\theta}\left(\mathbf{\hat{x}}_{t}, t\right) = \mathbf{K}_t\cdot \boldsymbol{\hat{\epsilon}}_{\theta}\left(\hat{\mathbf{x}}_{t}, t\right),
\end{equation}
where $\left(\cdot\right)$ denote the channel-wise multiplication. We now focus on finding a loss function $\mathcal{L}$, by minimizing which, we can efficiently reconstruct $\boldsymbol{\epsilon}_{\theta}\left(\mathbf{x}_{t}, t\right), \forall t\in[0,T]$: 
\begin{equation}
    \min \mathcal{L}(\boldsymbol{\epsilon}_{\theta}\left(\mathbf{x}_{t}, t\right), \boldsymbol{\tilde{\epsilon}}_{\theta}\left(\mathbf{\hat{x}}_{t}, t\right)).
    \label{eq:11}
\end{equation}

For brevity, in \cref{sec:ner}, we omit the input of the noise estimation network, using $\boldsymbol{\epsilon}_{t,i}$, $\boldsymbol{\hat{\epsilon}}_{t,i}$ and $\boldsymbol{\tilde{\epsilon}}_{t,i}$ to denote the estimation $\boldsymbol{\epsilon}_{t,i}\left(\mathbf{x_t,t}\right)$, $\boldsymbol{\hat{\epsilon}}_{t,i}\left(\mathbf{\hat{x}_t,t}\right)$ and $\boldsymbol{\tilde{\epsilon}}_{t,i}\left(\mathbf{\hat{x}_t,t}\right), i \in [0,C]$, respectively. 

While simply minimizing MSE between two images might cause blurriness, inspired by previous work in PTQ~\cite{finkelstein2019fighting}, we measure both the absolute error and the relative error in our loss function $\mathcal{L}$. For the convenience of computation and optimization, we use the inverse root quantization to noise ratio (rQNSR)~\cite{finkelstein2019fighting} to measure the signal's relative distortion caused by the quantization noise:
\begin{equation}
\text{rQNSR}\left(\boldsymbol{\hat{\epsilon}}_{t,i},\boldsymbol{\epsilon}_{t,i}\right) = \sqrt{\frac{\sum^{H}_{j}\sum^{W}_{k} \left(\boldsymbol{\hat{\epsilon}}_{t,i,j,k} - \boldsymbol{\epsilon}_{t,i,j,k}\right)^2}{\sum^{H}_{j}\sum^{W}_{k} \boldsymbol{\epsilon}_{t,i,j,k}^2}}, 
\label{eq:rqnsr}
\end{equation}
where $H$ and $W$ denote the height and weight of the noise estimation, respectively.
By taking the $\lambda_1$ as the weight of rQNSR penalty in $\mathcal{L}$ and $\lambda_2$ the regularization coefficient, we define the reconstruction loss for channel $i$ at timestep $t$ as:
\begin{equation}
\begin{aligned}
\mathcal{L}(\mathbf{K}_{t,i},\boldsymbol{\hat{\epsilon}}_{t,i},\boldsymbol{\epsilon}_{t,i}) =& 
\left(1-\lambda_1\right)\cdot \text{MSE}\left(\boldsymbol{\tilde{\epsilon}}_{t,i},\boldsymbol{\epsilon}_{t,i}\right)^2 + \lambda_1\cdot \text{rQNSR}\left(\boldsymbol{\tilde{\epsilon}}_{t,i},\boldsymbol{\epsilon}_{t,i}\right)^2 \\ &+ \lambda_2\cdot(\mathbf{K}_{t,i}-1)^2.
\label{eq:loss_c}
\end{aligned}
\end{equation}

Our next object is to find the optimal $\mathbf{K}_{t,i}$ that reduces the MSE and the rQNSR simultaneously. By expanding \cref{eq:loss_c}, we have:
\begin{equation}
\begin{aligned}
\mathcal{L}(\mathbf{K}_{t,i},\boldsymbol{\hat{\epsilon}}_{t,i},\boldsymbol{\epsilon}_{t,i}) 
=& \left(1-\lambda_1\right)\cdot \frac{1}{N}\sum^{H}_{j}\sum^{W}_{k} \left(\mathbf{K}_{t,i}\boldsymbol{\hat{\epsilon}}_{t,i,j,k}- \boldsymbol{\epsilon}_{t,i,j,k}\right)^2 \\
&+\lambda_1\cdot \frac{1}{N} \sum^{H}_{j}\sum^{W}_{k} \left(\frac{\mathbf{K}_{t,i}\hat{\boldsymbol{\epsilon}}_{t,i,j,k}-\boldsymbol{\epsilon}_{t,i,j,k}}{\boldsymbol{\epsilon}_{t,i,j,k}}\right)^2 +\lambda_2 \left(\mathbf{K}_{t,i} - 1\right)^2 \\
=& A_{t,i}\cdot \mathbf{K}_{t,i}^2 + B_{t,i}\cdot \mathbf{K}_{t,i} + D_{t,i},
\label{eq:loss_c_simplified}
\end{aligned}
\end{equation}

where
\begin{equation}
\begin{cases}
A_{t,i} = \frac{1-\lambda_1}{N} \sum^{H}_{j}\sum^{W}_{k} \hat{\boldsymbol{\epsilon}}_{t,i,j,k}^2 + \lambda_1 \sum^{H}_{j}\sum^{W}_{k}\frac{\hat{\boldsymbol{\epsilon}}_{t,i,j,k}^2}{\boldsymbol{\epsilon}_{t,i,j,k}^2}  + \lambda_2, \\
B_{t,i} = \frac{2\lambda_1-2}{N} \sum^{H}_{j}\sum^{W}_{k} \hat{\boldsymbol{\epsilon}}_{t,i,j,k}\boldsymbol{\epsilon}_{t,i,j,k}  - 2 \lambda_1 \sum^{H}_{j}\sum^{W}_{k} \frac{\hat{\boldsymbol{\epsilon}}_{t,i,j,k}}{\boldsymbol{\epsilon}_{t,i,j,k}} - 2\lambda_2, \\
D_{t,i}= \frac{1-\lambda_1}{N}\sum^{H}_{j}\sum^{W}_{k} \boldsymbol{\epsilon}_{t,i,j,k}^2 + \lambda_1 N  + \lambda_2,\\
N = C\cdot H\cdot W.
\end{cases}
\label{eq:coefficient_abd}
\end{equation}
$C$,$H$,$W$ denote the number of channels, height, and weight of the  estimation, respectively. 
The second derivative of the loss with respect to $\mathbf{K}_{t,i}$ is:
 
\begin{equation}
\frac{\partial^2\mathcal{L}_{t,i}}{\partial \mathbf{K}_{t,i}^2} = 2A_{t,i}.
\label{eq:second_order}
\end{equation}
By assigning both $\lambda_1$ and $\lambda_2$ as positive coefficients, we have $ A_{t,i} > 0 $. As a consequence, the reconstruction loss is convex, permitting us to minimize it with the closed-form solution which satisfies:
\begin{equation}
    \frac{\partial\mathcal{L}(\mathbf{K}_{t,i}) }{\partial \mathbf{K}_{t,i}} = 0 .
    \label{eq:first_order}
\end{equation} 
Plugging \cref{eq:coefficient_abd} into \cref{eq:first_order}, we have:
\begin{equation}
\begin{aligned}
    \mathbf{K}_{t,i} &= -\frac{B_{t,i}}{2A_{t,i}} =  \frac{ (1-\lambda_1)\sum^{H}_{j}\sum^{W}_{k} \hat{\boldsymbol{\epsilon}}_{t,i,j,k}\boldsymbol{\epsilon}_{t,i,j,k} + \lambda_1 N \sum^{H}_{j}\sum^{W}_{k}  \frac{\hat{\boldsymbol{\epsilon}}_{t,i,j,k}}{\boldsymbol{\epsilon}_{t,i,j,k}} + \lambda_2 N}{ (1-\lambda_1) \sum^{H}_{j}\sum^{W}_{k} \hat{\boldsymbol{\epsilon}}_{t,i,j,k}^2 + \lambda_1 N\sum^{H}_{j}\sum^{W}_{k} \frac{\hat{\boldsymbol{\epsilon}}_{t,i,j,k}^2}{\boldsymbol{\epsilon}_{t,i,j,k}^2} + \lambda_2 N}.
    \label{eq:c_without_mask}
\end{aligned} 
\end{equation}
 In practice, we pre-calculate $\mathbf{K}_{t,i}$ using a small batch of samples. Considering that larger values in $\mathbf{\epsilon}_{t,i}$ might indicate more change to $\mathbf{x}_{t,i}$~\cite{song2020denoising}, correcting the error in larger value other than that in smaller values might lead to more effective reconstruction. Therefore, when calculating $\mathbf{K}_{t,i}$, we only consider pixels whose values are larger than a threshold:
\begin{align}
    \tau_t =  k_{threshold}* \frac{1}{N}\sum^{C}_{i}\sum^{H}_{j}\sum^{W}_{k}|\boldsymbol{\epsilon}_{t,i,j,k}|, \tau \in \mathbb{R}^{T},
\end{align}
 with $k_{threshold}$ denoting the coefficient for threshold calculation. We use the element-wise mask $M \in \mathbb{R}^{T \times C \times H \times W }$ to filter pixels with large activation:
\begin{equation}
\begin{aligned}
    M_{t,i,j,k} = 
    \begin{cases} 
    1 & \text{if } \boldsymbol{\epsilon}_{t,i,j,k} > \tau_t, \\
    0 & \text{otherwise} .
    \end{cases}
    \label{eq:mask}
\end{aligned} 
\end{equation}
 
By replacing $\hat{\boldsymbol{\epsilon}}_{t,i}$ and $\boldsymbol{\epsilon}_{t,i}$ in \cref{eq:c_without_mask} with $\hat{\boldsymbol{\epsilon}}_{t,i} \cdot M$ and $\boldsymbol{\epsilon}_{t,i} \cdot M$, we calculate the reconstruction coefficient $K_{t,i}$ with following formula:

\begin{equation}
\begin{aligned}
    \mathbf{K}_{t,i} =  \frac{ (1-\lambda_1)\sum^{H}_{j}\sum^{W}_{k} \hat{\boldsymbol{\epsilon}}_{t,i,j,k}\boldsymbol{\epsilon}_{t,i,j,k}M_{t,i} + \lambda_1 N \sum^{H}_{j}\sum^{W}_{k}  \frac{\hat{\boldsymbol{\epsilon}}_{t,i,j,k}}{\boldsymbol{\epsilon}_{t,i,j,k}}M_{t,i} + \lambda_2 N}{ (1-\lambda_1) \sum^{H}_{j}\sum^{W}_{k} \hat{\boldsymbol{\epsilon}}_{t,i,j,k}^2M_{t,i}  + \lambda_1 N\sum^{H}_{j}\sum^{W}_{k} \frac{\hat{\boldsymbol{\epsilon}}_{t,i,j,k}^2}{\boldsymbol{\epsilon}_{t,i,j,k}^2}M_{t,i} + \lambda_2 N}.
\end{aligned}
\label{eq:c_with_mask}
\end{equation}
Both multiplication and division in \cref{eq:c_with_mask} are element-wise. After pre-calculating $\mathbf{K}_{t,i}$ for a small batch, this coefficient is directly used for estimation reconstruction during inference, as shown in \cref{fig:eccv_ppl}.

\subsection{Input Bias Correction}
\label{Sec:ibc}
Diffusion models are trained on the objective~\cite{ho2020denoising}:
\begin{equation}
L_{\text {simple }}(\theta):=\mathbb{E}_{t, \mathbf{x}_0, \boldsymbol{\epsilon}}\left[\left\|\boldsymbol{\epsilon}-\boldsymbol{\epsilon}_\theta\left(\sqrt{\bar{\alpha}_t} \mathbf{x}_0+\sqrt{1-\bar{\alpha}_t} \boldsymbol{\epsilon}, t\right)\right\|^2\right], 
\label{eq:objective}
\end{equation}
where $\theta$ denotes the model parameters, $\boldsymbol{\epsilon}$ denotes the target noise, $\boldsymbol{\epsilon}_{\theta}$ denotes the noise estimation network, $\Bar{\alpha}_t$ denotes the hyper-parameter and $\mathbf{x}_0$ denotes the input image. While in the training, the DMs are given the ground-truth input $\sqrt{\bar{\alpha}_t} \mathbf{x}_0+\sqrt{1-\bar{\alpha}_t} \boldsymbol{\epsilon}$, they never receive it as input during the inference, resulting in noise estimation error. Such training-inference discrepancy in auto-regressive generative models is named as exposure bias~\cite{schmidt-2019-generalization}. Recent works~\cite{li2024alleviating,ning2024elucidating,ning2023input} systematically analyze the exposure bias problem in full-precision diffusion models and succeed in improving the models' performance by addressing this problem. 

We argue that in diffusion models quantized with PTQ methods, the exposure bias problem is more severe, considering that the error introduced by quantization is far larger than the training-inference discrepancy studied in~\cite{li2024alleviating,ning2024elucidating,ning2023input,li2024on}. Thus it necessitates a solution to it. 

According to~\cite{ho2020denoising}, the training objective \cref{eq:objective} is simplified from:
\begin{equation}
\begin{split}
L_{t-1}-C = \mathbb{E}_{\mathbf{x}_0, \boldsymbol{\epsilon}}\left[\frac{1}{2 \sigma_t^2}\left\|\frac{1}{\sqrt{\alpha_t}}\left(\mathbf{x}_t\left(\mathbf{x}_0, \boldsymbol{\epsilon}\right)- \frac{\beta_t}{\sqrt{1-\bar{\alpha}_t}} \boldsymbol{\epsilon}\right)-\boldsymbol{\mu}_\theta\left(\mathbf{x}_t\left(\mathbf{x}_0, \boldsymbol{\epsilon}\right), t\right)\right\|^2\right],
\end{split}
\label{eq:reparameterization}
\end{equation}
where $C$ is a constant independent of $\theta$, and $\boldsymbol{\mu}_{\theta}\left(\mathbf{x}_t\left(\mathbf{x}_0,\boldsymbol{\epsilon}\right),t\right)$ is the model's prediction for the mean of $\mathbf{x}_{t-1}$.\cref{eq:reparameterization} indicates that the  $\boldsymbol{\mu}_{\theta}$ must predict $\frac{\beta_t}{\sqrt{1-\bar{\alpha}_t}} \boldsymbol{\epsilon}$ given $\mathbf{x}_t$~\cite{ho2020denoising}. Therefore, if the mean of $\mathbf{\hat{x}}_t$ is close to $\mathbf{x}_t$, low-precision DMs might be able to predict $\boldsymbol{\epsilon}_t$ better.

We propose Input Bias Correction (IBC) to correct the input discrepancy. We first calculate the average element-level bias in the deviated model input using a mini-batch of $S$ sample: 
\begin{equation}
    \mathbf{B}_{t,i,j,k} =\frac{1}{S} \sum^{S}_{s=0}\left(\mathbf{\hat{x}}_{t,s,i,j,k} - \mathbf{x}_{t,s,i,j,k}\right), \mathbf{B} \in \mathbb{R}^{T \times C \times H \times W},
\end{equation}
during the inference, the input discrepancy can be rectified with this correction:
\begin{equation}
    \tilde{\mathbf{x}}_{t-1} = \hat{\mathbf{x}}_{t-1} - \mathbf{B}_{t-1}.
\end{equation}

We further compare the IBC with the estimation bias correction strategy (solely correct the mean of $\boldsymbol{\hat{\epsilon}}_t$), and find that IBC performs better, which indicates that directly addressing the exposure bias problem in low-precision DMs is an effective method (see \cref{sec:comparison_ibc}).

\begin{algorithm}[t]
\caption{Timestep-Aware Correction}
\begin{algorithmic}

\State \textbf{Pre-calculation:} 
\State \textbf{Input: } Full-precision diffusion model $\boldsymbol{\epsilon_t}$ and its quantized version $\boldsymbol{\hat{\epsilon}_t}$ 
\State \textbf{Output:} Reconstruction Coefficient $\mathbf{K}_{t,i}$ and Bias Corrector $\mathbf{B}_{t,i,j,k}$
\For{$t = T$ to $0$}
    \State Collect model output $\boldsymbol{\epsilon_{\theta}\left(\mathbf{x}_t,t\right)}$  and $\boldsymbol{\hat{\epsilon}_{\theta}\left(\mathbf{\hat{x}}_t,t\right)}$
    \State Calculate the reconstruction coefficient $\mathbf{K}_{t,i}$ 
    \State Calculate the element-level bias $\mathbf{B}_{t,i,j,k}$
    \State Correct $\boldsymbol{\hat{\epsilon}_{\theta}\left(\mathbf{\hat{x}}_t,t\right)}$ and $\mathbf{\hat{x}}_{t-1}$ with $\mathbf{K}_{t,i}$ and $\mathbf{B}_{t,i,j,k}$
    \State Save $\mathbf{K}_{t,i}$ and $\Delta\mathbf{x}_{t-1}$ for inference 
\EndFor
\\
\State \textbf{Inference:} 
\State \textbf{Input: } Quantized model $\boldsymbol{\hat{\epsilon}_t}$, coefficient $\mathbf{K}_{t,i}$ and corrector $\mathbf{B}_{t,i,j,k}$
\State \textbf{Output:} Corrected Output $\mathbf{\tilde{x}}_0$
\For{$t = T$ to $0$}
    \State Correct input $\mathbf{\hat{x}}_{t,i,j,k}$ with $\mathbf{B}_{t,i,j,k}$
    \State Estimate noise with corrected input $\mathbf{\tilde{x}}_{t,i,j,k} $
    \State Reconstruct noise estimation $\boldsymbol{{\epsilon}}_{t,i}$ with $\mathbf{K}_{t,i}$ and $\boldsymbol{\hat{\epsilon}}_{t,i}$
\EndFor
\end{algorithmic}
\label{alg:final}
\end{algorithm}

The overall timestep-aware correction pipeline, including the pre-process and the inference process, is described in Algorithm~\ref{alg:final}. During inference, the corrected sampling process for low-precision DMs can be formulated as:
\begin{equation}
\tilde{\mathbf{x}}_{t-1}=\frac{1}{\sqrt{\alpha_{t}}}\left(\mathbf{\hat{x}}_{t}-\frac{\beta_{t}}{\sqrt{1-\bar{\alpha}_{t}}}\mathbf{K}_t\boldsymbol{\hat{\epsilon}}_{\theta}\left(\mathbf{\hat{x}}_{t}-\mathbf{B}_{t}, t\right) - \mathbf{B}_{t}\right) + \mathbf{z}, \mathbf{z} \sim \mathcal{N}(\mathbf{0}, \mathbf{I}).
\label{eq:integrated_correction}
\end{equation}

\section{Experiments}
\label{sec:experiments}

\subsection{Experiments Setup}
Following previous work for diffusion quantization~\cite{li2023q,shang2023post,he2023ptqd,he2024efficientdm}, we evaluate our method on both unconditional and conditional generation. For unconditional generation, we conduct experiments with quantized DDIM on CIFAR-10~\cite{krizhevsky2009learning}, LDM-8 on LSUN-Church~\cite{yu15lsun} and LDM-4 on LSUN-Bedroom. For conditional generation, we evaluate our method with Stable-Diffusion v1.4~\cite{nokey} on text-to-image task. Notably, even with unseen prompts, TAC-Diffusion still achieves considerable improvement in generated images (see \cref{sec:stable_diffusion}). In experiments of 3-bit and 2-bit weight quantization, we compare our model with the state-of-the-art PTQ methods, including PTQ4DM~\cite{shang2023post}, Q-Diffusion~\cite{li2023q}, and PTQD~\cite{he2023ptqd}, implementing their result by rerunning their official codes. The primary metric for diffusion models is Fréchet Inception Distance (FID, the lower the better)~\cite{heusel2017gans}, we report it for all experiments. We also evaluate the Inception Score (IS, the higher the better)~\cite{salimans2016improved} for experiments on CIFAR-10. To ensure consistency with previous works, all results are obtained by sampling 50,000 images. To further investigate the portability of the proposed method, we extend it to another fast sampling method, DPM-solver$++$~\cite{lu2023dpmsolver} (see \cref{sec:dpm-solver}).

\subsection{Main Results}

We conduct extensive experiments on CIFAR-10~\cite{krizhevsky2009learning} using a 100-step DDIM, across four quantization precisions: W8A8, W4A8, W3A8, and W3A6. Quantitative results are shown in \cref{tab:cifar} (see the \cref{sec:cifar_qualitative} for corresponding qualitative results). We observe that, across all precision levels, TAC-Diffusion achieves better image fidelity and diversity, even outperforming the QAT method EfficientDM~\cite{he2024efficientdm} under W8A8 quantization. At lower precisions, such as W3A8 and W3A6, TAC-Diffusion demonstrates robust performance, maintaining a relatively low FID of 9.55 and 31.77, respectively, thus narrowing the performance gap between a quantized model and its full-precision counterpart.

\begin{table}[!ht]
\centering
\footnotesize
\caption{Quantization results for unconditional image generation with 100 steps DDIM on CIFAR-10 (32 $\times$ 32).}
\label{tab:cifar}
\begin{tabular}{llcccc}
\toprule
\textbf{Model} & \textbf{Method} & \textbf{Bits(W/A)} & \textbf{Size(Mb)} & \textbf{IS}${\uparrow}$ & \textbf{FID}${\downarrow}$ \\ 
\midrule
\multirow{21}{*}{\parbox{3cm}{DDIM \\ (steps = 100 \\ eta = 0.0 )}} & FP &  32/32 & 143.2 & 9.12 & 4.22 \\
\cline{2-6}
&PTQ4DM~\cite{shang2023post}  & 8/8 &  34.26 & 9.31 & 10.55 \\
&Q-diffusion~\cite{li2023q}  & 8/8 &   35.8 & 9.48 & 3.75 \\
&TDQ~\cite{so2023temporal}   & 8/8 &  34.30 & 8.85 & 5.99 \\
&EDA-DM~\cite{liu2024enhanced}  & 8/8 & -- & 9.40 & 3.72 \\
&TFMQ-DM~\cite{huang2023tfmq}  & 8/8 & -- & 9.07 & 4.24 \\
&EfficientDM (QAT)~\cite{he2024efficientdm}  & 8/8 & 34.30 & 9.38 & 3.75 \\
&Ours & 8/8 &  35.8 & \textbf{9.49} & \textbf{3.68} \\

\cline{2-6}
&PTQ4DM~\cite{shang2023post}  & 4/8 &  17.22 & 7.92 & 37.79 \\
&Q-diffusion~\cite{li2023q}  & 4/8 &  17.31 & 9.12 & 4.93 \\
&EDA-DM~\cite{liu2024enhanced}  & 4/8 & -- & 9.29 & 4.16 \\
&TFMQ-DM~\cite{huang2023tfmq}  & 4/8 & -- & 9.13 & 4.78 \\
&EfficientDM (QAT)~\cite{he2024efficientdm}  & 4/8 & 17.26 & \textbf{9.41 }& \textbf{3.80} \\
&Ours & 4/8 & 17.31 & 9.15 & 4.89 \\

\cline{2-6}
&PTQ4DM~\cite{shang2023post} & 3/8 & 13.4 & 4.21 & 115.4 \\
&Q-diffusion~\cite{li2023q}  & 3/8 & 13.4 & 8.53 & 17.31 \\
&Ours & 3/8 & 13.4 & \textbf{8.86}& \textbf{9.55} \\

\cline{2-6}
&PTQ4DM~\cite{shang2023post} & 3/6 & 13.4 & 4.07 & 124.97 \\ 
&Q-diffusion~\cite{li2023q} & 3/6 & 13.4 & \textbf{8.54} & 33.21 \\
&Ours & 3/6 & 13.4 & 8.27 & \textbf{31.77} \\
\bottomrule
\end{tabular} 
\vspace{-0.4cm}
\end{table}

Evaluations using LDM-4 and LDM-8 on the LSUN-Church and LSUN-Bedroom datasets are detailed in \cref{tab:lsun}. The quantitative outcomes for these high-resolution datasets reflect the performance improvements previously observed in experiments with lower resolutions (see \cref{tab:cifar}). Remarkably, TAC-Diffusion significantly boosts the performance of W3A8/W2A8 quantized LDM-8 models, achieving FID reductions of 2.03 and 1.69, respectively, compared to the baseline reported in ~\cite{li2023q}, thereby pushing the limits of post-training quantization (PTQ) for diffusion models to 2-bit precision for the first time.

\begin{table}[!ht]
\centering
\caption{Quantization results for unconditional image generation with 500 steps LDM-8 on LSUN-Churches (256$\times$256) and 200 steps LDM-4 on LSUN-Bedrooms (256$\times$256).}
\begin{tabular}{llcccc}
\toprule
\textbf{Model} & \textbf{Method} & \textbf{Bits (W/A)} & \textbf{Size (Mb)} & \textbf{TBops} &  \textbf{FID}${\downarrow}$  \\ 
\midrule
\multirow{16}{*}{\parbox{3cm}{LDM-8 \\ (steps = 500 \\ eta = 0.0 )}} & Full-precision & 32/32 & 1179.9 & 22.17 & 4.06 \\
\cline{2-6}
& PTQD~\cite{he2023ptqd} & 8/8 & 295.0 & 2.68 & 10.76 \\
& QuEST~\cite{wang2024quest} & 8/8 & 330.6  & -- & 6.55 \\
& TFMQ-DM~\cite{huang2023tfmq} & 8/8 & 295.0 & -- & 4.01 \\
& EDA-DM\cite{liu2024enhanced} & 8/8 & -- & -- & 3.83 \\
& Q-diffusion~\cite{li2023q} & 8/8 & 295.0 & 2.68 & 3.65 \\
&Ours & 8/8 & 295.0 & 2.68 & \textbf{3.37} \\
\cline{2-6}

& PTQD~\cite{he2023ptqd} & 4/8 & 147.5 & 1.34 & 8.41 \\
& QuEST~\cite{wang2024quest} & 4/8 & 189.9 & -- & 7.33 \\
& TFMQ-DM~\cite{huang2023tfmq} & 4/8 & 147.5 & -- & 4.14 \\
& EDA-DM\cite{liu2024enhanced} & 4/8 & -- & -- & 4.01 \\
& Q-diffusion~\cite{li2023q} & 4/8 & 147.5 & 1.34 & 4.12 \\
&Ours & 4/8 & 147.5 & 1.34 & \textbf{3.81} \\
\cline{2-6}

& PTQD~\cite{he2023ptqd} & 3/8 & 110.6 & 1.01 & 12.68 \\
& Q-diffusion~\cite{li2023q} & 3/8 & 110.6 & 1.01 & 9.80 \\
&Ours & 3/8 & 110.6 & 1.01 & \textbf{7.78} \\
\hline
\multirow{9}{*}{\parbox{3cm}{LDM-4 \\ (steps = 200 \\ eta = 1.0 )}} &Full-precision & 32/32 & 1096.2 & 107.17 & 2.98 \\
\cline{2-6}
& PTQD~\cite{he2023ptqd} & 4/8 & 137.0 & 6.48 & 5.94 \\
& Q-diffusion~\cite{li2023q} & 4/8 & 137.0 & 6.48 & 5.32 \\
&Ours & 4/8 & 137.0 &  6.48&  \textbf{4.94}\\
\cline{2-6}
& PTQD~\cite{he2023ptqd} & 3/8 & 102.75 & 4.86& 6.46 \\
& Q-diffusion~\cite{li2023q} & 3/8 & 102.75 & 4.86 & 7.17 \\
&Ours & 3/8 & 102.75 & 4.86 & \textbf{5.14}  \\
\cline{2-6}
& Q-Diffusion & 2/8 & 68.5 & 2.43 &  9.48 \\
&Ours & 2/8 & 68.5 & 2.43 & \textbf{7.79}  \\
\bottomrule 
\label{tab:lsun}
\end{tabular}
\vspace{-0.4cm}
\end{table}

\begin{figure}[!ht]
    \centering  

    \subcaptionbox{FP32}{
            \includegraphics[width=0.16\linewidth]{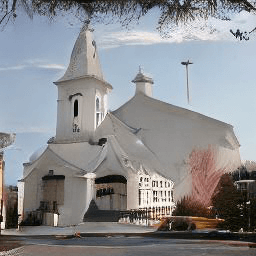}
            \includegraphics[width=0.16\linewidth]{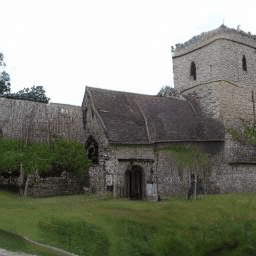}
            \includegraphics[width=0.16\linewidth]{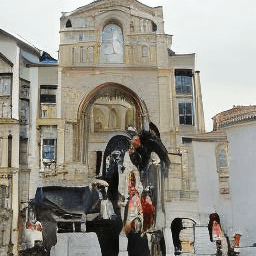}
            \includegraphics[width=0.16\linewidth]{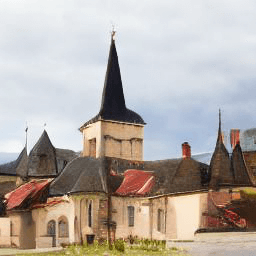}
            \includegraphics[width=0.16\linewidth]{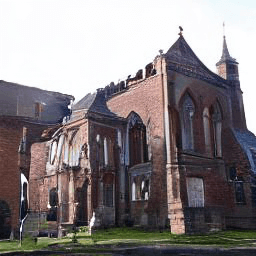}
            \includegraphics[width=0.16\linewidth]{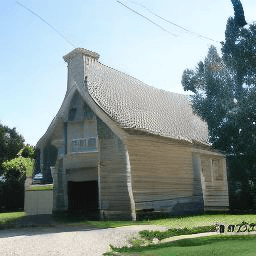}
    }
    \subcaptionbox{TAC-Diffusion W3A8}{
            \includegraphics[width=0.16\linewidth]{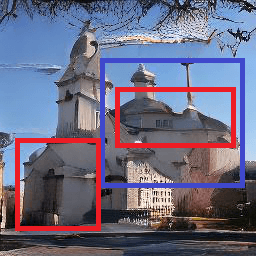}
            \includegraphics[width=0.16\linewidth]{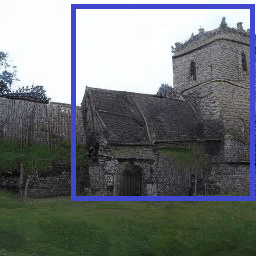}
            \includegraphics[width=0.16\linewidth]{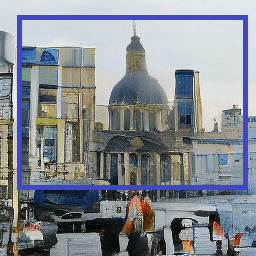}
            \includegraphics[width=0.16\linewidth]{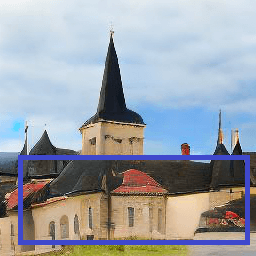}
            \includegraphics[width=0.16\linewidth]{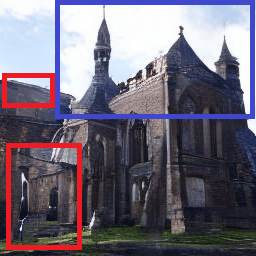}
            \includegraphics[width=0.16\linewidth]{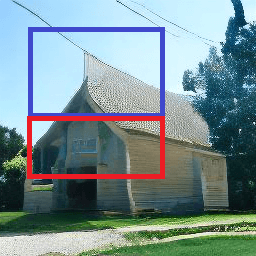}
    }
    \subcaptionbox{Q-Diffusion W3A8}{
            \includegraphics[width=0.16\linewidth]{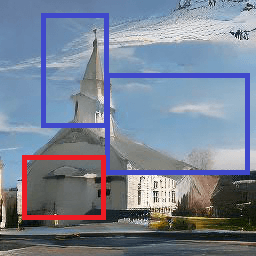}
            \includegraphics[width=0.16\linewidth]{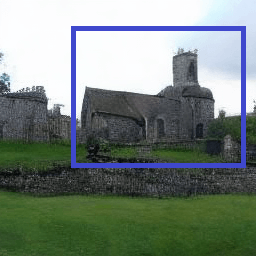}
            \includegraphics[width=0.16\linewidth]{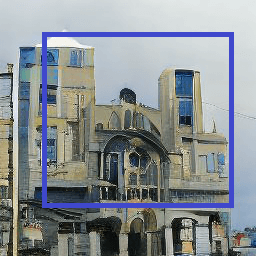}
            \includegraphics[width=0.16\linewidth]{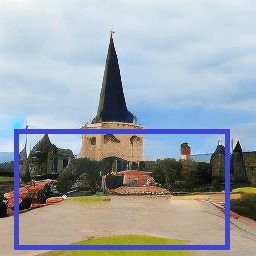}
            \includegraphics[width=0.16\linewidth]{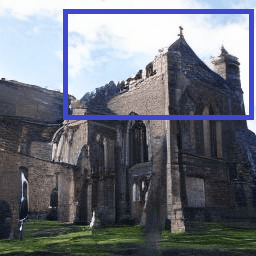}
            \includegraphics[width=0.16\linewidth]{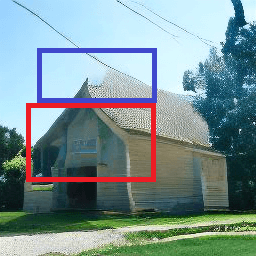}
    }
    \subcaptionbox{PTQD W3A8}{
            \includegraphics[width=0.16\linewidth]{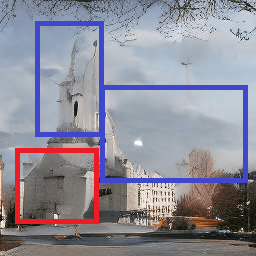}
            \includegraphics[width=0.16\linewidth]{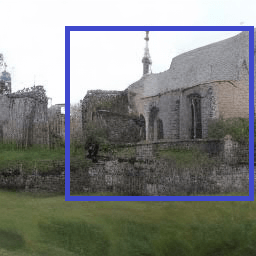}
            \includegraphics[width=0.16\linewidth]{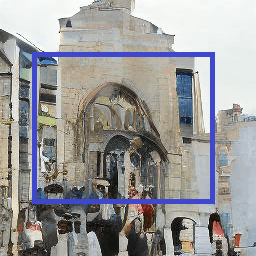}
            \includegraphics[width=0.16\linewidth]{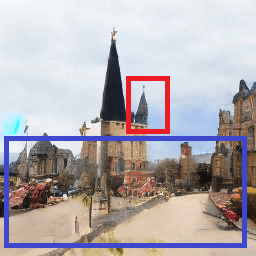}
            \includegraphics[width=0.16\linewidth]{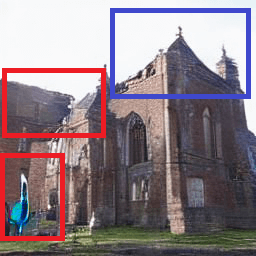}
            \includegraphics[width=0.16\linewidth]{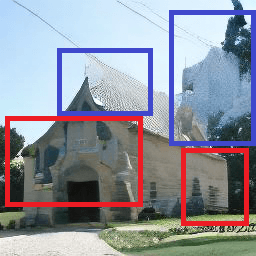}
    }
  \caption{ 256 × 256 unconditional image generation results with W3A8 LDM-8~\cite{rombach2022high}, on LSUN-Church dataset. Red boxes highlight areas where our model preserves intricate details more effectively. Blue boxes show regions where our model maintains structural accuracy, closely resembling the full-precision model's output.}
  \label{fig:church16}
\vspace{-2cm} 
\end{figure}

\begin{figure}[!ht]
    \centering
    \subcaptionbox{FP32}{
        \includegraphics[width = 0.16\linewidth]{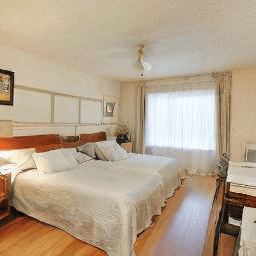}
        \includegraphics[width = 0.16\linewidth]{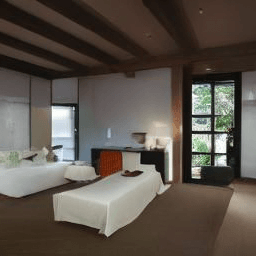}
        \includegraphics[width = 0.16\linewidth]{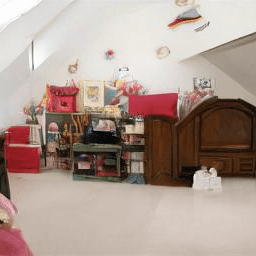}
        \includegraphics[width = 0.16\linewidth]{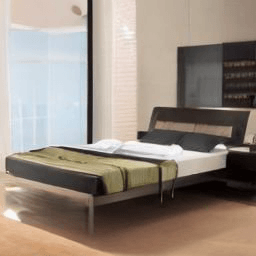}
        \includegraphics[width = 0.16\linewidth]{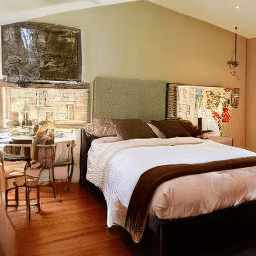}   
        \includegraphics[width = 0.16\linewidth]{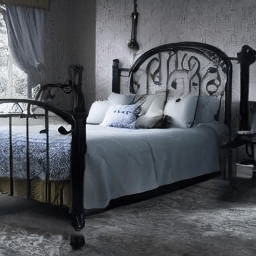}
    }
    \subcaptionbox{TAC-Diffusion W3A8}{
        \includegraphics[width = 0.16\linewidth]{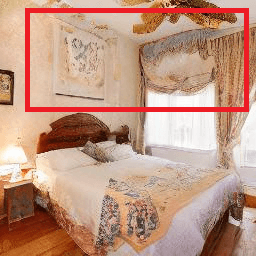}
        \includegraphics[width = 0.16\linewidth]{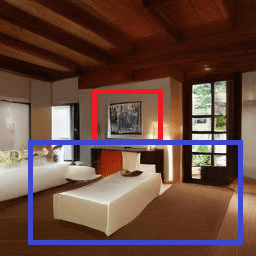}
        \includegraphics[width = 0.16\linewidth]{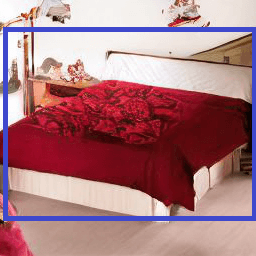}
        \includegraphics[width = 0.16\linewidth]{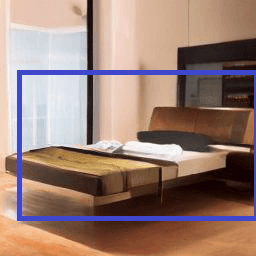}
        \includegraphics[width = 0.16\linewidth]{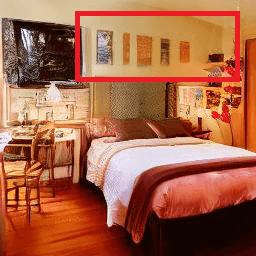}   
        \includegraphics[width = 0.16\linewidth]{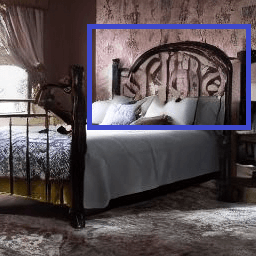}
    }
    \subcaptionbox{Q-Diffusion W3A8}{
        \includegraphics[width = 0.16\linewidth]{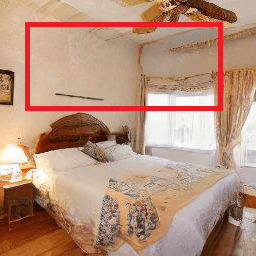}
        \includegraphics[width = 0.16\linewidth]{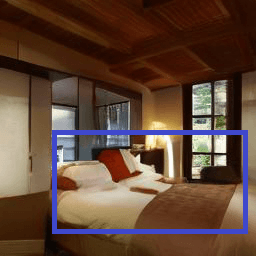}
        \includegraphics[width = 0.16\linewidth]{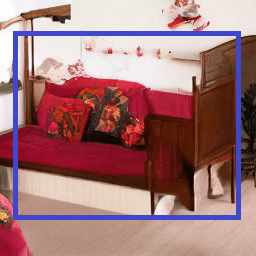}
        \includegraphics[width = 0.16\linewidth]{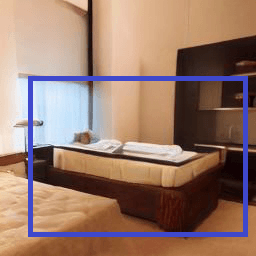}
        \includegraphics[width = 0.16\linewidth]{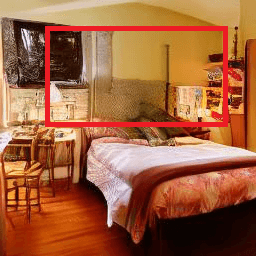}   
        \includegraphics[width = 0.16\linewidth]{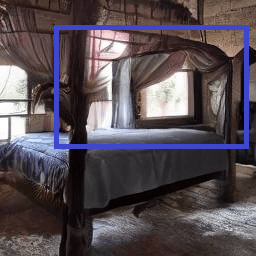}
    }
    \caption{Comparative analysis of 256 $\times$ 256 unconditional images generation with 200 steps latent diffusion model, on LSUN-Bedroom dataset. Red boxes showcase our model's enhanced detail preservation, and blue boxes emphasize its superior structural accuracy.}
    \label{fig:bedroom}
      \vspace{-0.4cm}
\end{figure}

The corresponding qualitative results on LSUN-Church and LSUN-Bedroom are presented in \cref{fig:church16} and the \cref{fig:bedroom}. We compare TAC-Diffusion with the full-precision model, Q-Diffusion~\cite{li2023q}. These visualization demonstrates TAC-Diffusion's superiority in preserving detailed textures and structural integrity, which are highlighted by red and blue boxes, respectively. We argue that this success is due to TAC-Diffusion's ability to reduce errors during both the early and later stages, when structure and details begin to emerge~\cite{fang2024structural}.

\subsection{Ablation Study}
\label{sec:ablation_study}

\begin{table}[!ht]
\centering
\caption{Ablation study of each component in our method, with W3A8 100 steps DDIM on CIFAR-10 (32 $\times$ 32).}
\begin{tabular}{lcll}
\toprule
\textbf{Method} & \textbf{Bits(W/A)}& \textbf{IS}${\uparrow}$  &  \textbf{FID}${\downarrow}$  \\ 
\hline
Baseline~\cite{li2023q}  & 3/8 &  8.53 & 17.31\\
+IBC+ Timestep-Aware & 3/8 & 8.60 \greentext{(+0.07)}&  16.16 \bluetext{(-1.15)}\\
+NER + IBC & 3/8 & 8.67 \greentext{(+0.14)} &  13.42 \bluetext{(-3.89)}\\
+NER + IBC+ Timestep-Aware (Ours) & 3/8 & \textbf{8.86} \greentext{(+0.33)} & \textbf{9.55} \bluetext{(-7.76)} \\
\bottomrule
\label{tab:ablation}
\end{tabular} 
\end{table}

To evaluate the individual contributions of each component in our proposed methods, we conduct a thorough ablation study using the CIFAR-10 dataset with a W3A8 100-step DDIM sampler. The results are documented in \cref{tab:ablation}, where 'IBC' represents Input Bias Correction, 'NER' stands for Noise Estimation Reconstruction, and 'Timestep-Aware' indicates that corrections were applied across all timesteps, as opposed to methods where corrections were only applied at the first timestep. We initiated our evaluation with the strong Q-Diffusion baseline~\cite{li2023q}, to evaluate the adaptability of our methods. The findings from our ablation study are summarized as follows:

\textbf{(a)} IBC effectively improves model's noise estimation accuracy, as evidenced by a 1.15 decrease in FID, underscoring our previous discussion on exposure bias in \cref{Sec:ibc}.
\textbf{(b)} NER successfully reconstructs the noise estimation from the corrupted noise estimation, leading to a notable 7.76 decrease in FID. This also verifies our assumption in \cref{sec:ner} that we can partially reconstruct the original noise estimation from the corrupted one with a appropriate scale factor.
\textbf{(c)} Applying a singular correction at the initial timestep yields a 3.89 FID reduction. This is attributed to the reduction of the error at the beginning greatly helps alleviate the error introduced in subsequent timesteps. thereby mitigating the error at minimal cost.
\textbf{(d)} Timestep-Aware Correction, by actively minimizing errors at each timestep, substantially lowers cumulative errors in the final output, leading to a 3.87 decrease in FID compared to the singular correction strategy.

\section{Conclusion}
\label{sec:conclusion}
In this paper, we present TAC-Diffusion, a Timestep-Aware Correction method that dynamically rectifies the quantization error in the denoising process of diffusion models. First, we assume that reducing the input discrepancy for each timestep will also minimize the error accumulated on the final output. Next, We propose to decompose the input discrepancy into two parts and rectify each one separately. Extensive experiments demonstrate that when employing our proposed method, diffusion models quantized to low precision can generate images that are on par with or even better than the full-precision model.

%
%

\bibliographystyle{splncs04}
\bibliography{egbib}

\begin{thebibliography}{10}
\providecommand{\url}[1]{\texttt{#1}}
\providecommand{\urlprefix}{URL }
\providecommand{\doi}[1]{https://doi.org/#1}

\bibitem{Bai_2023_ICCV}
Bai, S., Chen, J., Shen, X., Qian, Y., Liu, Y.: Unified data-free compression: Pruning and quantization without fine-tuning. In: Proceedings of the IEEE/CVF International Conference on Computer Vision (ICCV). pp. 5876--5885 (October 2023)

\bibitem{banner2019post}
Banner, R., Nahshan, Y., Soudry, D.: Post training 4-bit quantization of convolutional networks for rapid-deployment. Advances in Neural Information Processing Systems  \textbf{32} (2019)

\bibitem{bao2022analyticdpm}
Bao, F., Li, C., Zhu, J., Zhang, B.: Analytic-dpm: an analytic estimate of the optimal reverse variance in diffusion probabilistic models (2022)

\bibitem{bondarenko2021understanding}
Bondarenko, Y., Nagel, M., Blankevoort, T.: Understanding and overcoming the challenges of efficient transformer quantization. arXiv preprint arXiv:2109.12948  (2021)

\bibitem{chen2023data}
Chen, J., Bai, S., Huang, T., Wang, M., Tian, G., Liu, Y.: Data-free quantization via mixed-precision compensation without fine-tuning. Pattern Recognition p. 109780 (2023)

\bibitem{dhariwal2021diffusion}
Dhariwal, P., Nichol, A.: Diffusion models beat gans on image synthesis. Advances in neural information processing systems  \textbf{34},  8780--8794 (2021)

\bibitem{esser2020learned}
Esser, S.K., McKinstry, J.L., Bablani, D., Appuswamy, R., Modha, D.S.: Learned step size quantization (2020)

\bibitem{fang2024structural}
Fang, G., Ma, X., Wang, X.: Structural pruning for diffusion models. Advances in neural information processing systems  \textbf{36} (2024)

\bibitem{finkelstein2019fighting}
Finkelstein, A., Almog, U., Grobman, M.: Fighting quantization bias with bias. arXiv preprint arXiv:1906.03193  (2019)

\bibitem{gholami2022survey}
Gholami, A., Kim, S., Dong, Z., Yao, Z., Mahoney, M.W., Keutzer, K.: A survey of quantization methods for efficient neural network inference. In: Low-Power Computer Vision, pp. 291--326. Chapman and Hall/CRC (2022)

\bibitem{gu2022vector}
Gu, S., Chen, D., Bao, J., Wen, F., Zhang, B., Chen, D., Yuan, L., Guo, B.: Vector quantized diffusion model for text-to-image synthesis. In: Proceedings of the IEEE/CVF Conference on Computer Vision and Pattern Recognition. pp. 10696--10706 (2022)

\bibitem{han2015deep}
Han, S., Mao, H., Dally, W.J.: Deep compression: Compressing deep neural networks with pruning, trained quantization and huffman coding. arXiv preprint arXiv:1510.00149  (2015)

\bibitem{he2016deep}
He, K., Zhang, X., Ren, S., Sun, J.: Deep residual learning for image recognition. In: Proceedings of the IEEE conference on computer vision and pattern recognition. pp. 770--778 (2016)

\bibitem{he2024efficientdm}
He, Y., Liu, J., Wu, W., Zhou, H., Zhuang, B.: Efficient{DM}: Efficient quantization-aware fine-tuning of low-bit diffusion models. In: The Twelfth International Conference on Learning Representations (2024), \url{https://openreview.net/forum?id=UmMa3UNDAz}

\bibitem{he2023ptqd}
He, Y., Liu, L., Liu, J., Wu, W., Zhou, H., Zhuang, B.: {PTQD}: Accurate post-training quantization for diffusion models. In: Thirty-seventh Conference on Neural Information Processing Systems (2023), \url{https://openreview.net/forum?id=Y3g1PV5R9l}

\bibitem{heusel2017gans}
Heusel, M., Ramsauer, H., Unterthiner, T., Nessler, B., Hochreiter, S.: Gans trained by a two time-scale update rule converge to a local nash equilibrium. Advances in neural information processing systems  \textbf{30} (2017)

\bibitem{ho2020denoising}
Ho, J., Jain, A., Abbeel, P.: Denoising diffusion probabilistic models. Advances in neural information processing systems  \textbf{33},  6840--6851 (2020)

\bibitem{howard2017mobilenets}
Howard, A.G., Zhu, M., Chen, B., Kalenichenko, D., Wang, W., Weyand, T., Andreetto, M., Adam, H.: Mobilenets: Efficient convolutional neural networks for mobile vision applications. arXiv preprint arXiv:1704.04861  (2017)

\bibitem{huang2023tfmq}
Huang, Y., Gong, R., Liu, J., Chen, T., Liu, X.: Tfmq-dm: Temporal feature maintenance quantization for diffusion models. In: Conference on Computer Vision and Pattern Recognition (CVPR) (2024)

\bibitem{hubara2020improving}
Hubara, I., Nahshan, Y., Hanani, Y., Banner, R., Soudry, D.: Improving post training neural quantization: Layer-wise calibration and integer programming. arXiv preprint arXiv:2006.10518  (2020)

\bibitem{jacob2018quantization}
Jacob, B., Kligys, S., Chen, B., Zhu, M., Tang, M., Howard, A., Adam, H., Kalenichenko, D.: Quantization and training of neural networks for efficient integer-arithmetic-only inference. In: Proceedings of the IEEE conference on computer vision and pattern recognition. pp. 2704--2713 (2018)

\bibitem{krizhevsky2009learning}
Krizhevsky, A., Hinton, G.: Learning multiple layers of features from tiny images. Tech. Rep.~0, University of Toronto, Toronto, Ontario (2009), \url{https://www.cs.toronto.edu/~kriz/learning-features-2009-TR.pdf}

\bibitem{li2024alleviating}
Li, M., Qu, T., Yao, R., Sun, W., Moens, M.F.: Alleviating exposure bias in diffusion models through sampling with shifted time steps. In: The Twelfth International Conference on Learning Representations (2024), \url{https://openreview.net/forum?id=ZSD3MloKe6}

\bibitem{li2023q}
Li, X., Liu, Y., Lian, L., Yang, H., Dong, Z., Kang, D., Zhang, S., Keutzer, K.: Q-diffusion: Quantizing diffusion models. In: Proceedings of the IEEE/CVF International Conference on Computer Vision. pp. 17535--17545 (2023)

\bibitem{li2024on}
Li, Y., van~der Schaar, M.: On error propagation of diffusion models. In: The Twelfth International Conference on Learning Representations (2024), \url{https://openreview.net/forum?id=RtAct1E2zS}

\bibitem{li2023qdm}
Li, Y., Xu, S., Cao, X., Sun, X., Zhang, B.: Q-{DM}: An efficient low-bit quantized diffusion model. In: Thirty-seventh Conference on Neural Information Processing Systems (2023), \url{https://openreview.net/forum?id=sFGkL5BsPi}

\bibitem{li2021brecq}
Li, Y., Gong, R., Tan, X., Yang, Y., Hu, P., Zhang, Q., Yu, F., Wang, W., Gu, S.: Brecq: Pushing the limit of post-training quantization by block reconstruction. arXiv preprint arXiv:2102.05426  (2021)

\bibitem{liu2024enhanced}
Liu, X., Li, Z., Xiao, J., Gu, Q.: Enhanced distribution alignment for post-training quantization of diffusion models. arXiv preprint arXiv:2401.04585  (2024)

\bibitem{liu2021post}
Liu, Z., Wang, Y., Han, K., Zhang, W., Ma, S., Gao, W.: Post-training quantization for vision transformer. Advances in Neural Information Processing Systems  \textbf{34},  28092--28103 (2021)

\bibitem{lu2022dpm}
Lu, C., Zhou, Y., Bao, F., Chen, J., Li, C., Zhu, J.: Dpm-solver: A fast ode solver for diffusion probabilistic model sampling in around 10 steps. arXiv preprint arXiv:2206.00927  (2022)

\bibitem{lu2023dpmsolver}
Lu, C., Zhou, Y., Bao, F., Chen, J., Li, C., Zhu, J.: Dpm-solver++: Fast solver for guided sampling of diffusion probabilistic models (2023)

\bibitem{nagel2020up}
Nagel, M., Amjad, R.A., Van~Baalen, M., Louizos, C., Blankevoort, T.: Up or down? adaptive rounding for post-training quantization. In: International Conference on Machine Learning. pp. 7197--7206. PMLR (2020)

\bibitem{nagel2019data}
Nagel, M., Baalen, M.v., Blankevoort, T., Welling, M.: Data-free quantization through weight equalization and bias correction. In: Proceedings of the IEEE/CVF International Conference on Computer Vision. pp. 1325--1334 (2019)

\bibitem{nagel2021white}
Nagel, M., Fournarakis, M., Amjad, R.A., Bondarenko, Y., Van~Baalen, M., Blankevoort, T.: A white paper on neural network quantization. arXiv preprint arXiv:2106.08295  (2021)

\bibitem{nichol2021improved}
Nichol, A.Q., Dhariwal, P.: Improved denoising diffusion probabilistic models. In: International Conference on Machine Learning. pp. 8162--8171. PMLR (2021)

\bibitem{ning2024elucidating}
Ning, M., Li, M., Su, J., Salah, A.A., Ertugrul, I.O.: Elucidating the exposure bias in diffusion models. In: The Twelfth International Conference on Learning Representations (2024), \url{https://openreview.net/forum?id=xEJMoj1SpX}

\bibitem{ning2023input}
Ning, M., Sangineto, E., Porrello, A., Calderara, S., Cucchiara, R.: Input perturbation reduces exposure bias in diffusion models. arXiv preprint arXiv:2301.11706  (2023)

\bibitem{polino2018model}
Polino, A., Pascanu, R., Alistarh, D.: Model compression via distillation and quantization. arXiv preprint arXiv:1802.05668  (2018)

\bibitem{rombach2022high}
Rombach, R., Blattmann, A., Lorenz, D., Esser, P., Ommer, B.: High-resolution image synthesis with latent diffusion models. In: Proceedings of the IEEE/CVF conference on computer vision and pattern recognition. pp. 10684--10695 (2022)

\bibitem{nokey}
Rombach, R., Blattmann, A., Lorenz, D., Esser, P., Ommer, B.: High-resolution image synthesis with latent diffusion models. In: Proceedings of the IEEE Conference on Computer Vision and Pattern Recognition (CVPR) (2022), \url{https://github.com/CompVis/latent-diffusionhttps://arxiv.org/abs/2112.10752}

\bibitem{ronneberger2015u}
Ronneberger, O., Fischer, P., Brox, T.: U-net: Convolutional networks for biomedical image segmentation. In: Medical Image Computing and Computer-Assisted Intervention--MICCAI 2015: 18th International Conference, Munich, Germany, October 5-9, 2015, Proceedings, Part III 18. pp. 234--241. Springer (2015)

\bibitem{ruiz2023dreambooth}
Ruiz, N., Li, Y., Jampani, V., Pritch, Y., Rubinstein, M., Aberman, K.: Dreambooth: Fine tuning text-to-image diffusion models for subject-driven generation. In: Proceedings of the IEEE/CVF Conference on Computer Vision and Pattern Recognition. pp. 22500--22510 (2023)

\bibitem{saharia2022palette}
Saharia, C., Chan, W., Chang, H., Lee, C., Ho, J., Salimans, T., Fleet, D., Norouzi, M.: Palette: Image-to-image diffusion models. In: ACM SIGGRAPH 2022 Conference Proceedings. pp. 1--10 (2022)

\bibitem{saharia2022photorealistic}
Saharia, C., Chan, W., Saxena, S., Li, L., Whang, J., Denton, E.L., Ghasemipour, K., Gontijo~Lopes, R., Karagol~Ayan, B., Salimans, T., et~al.: Photorealistic text-to-image diffusion models with deep language understanding. Advances in Neural Information Processing Systems  \textbf{35},  36479--36494 (2022)

\bibitem{salimans2016improved}
Salimans, T., Goodfellow, I., Zaremba, W., Cheung, V., Radford, A., Chen, X.: Improved techniques for training gans. Advances in neural information processing systems  \textbf{29} (2016)

\bibitem{schmidt-2019-generalization}
Schmidt, F.: Generalization in generation: A closer look at exposure bias. In: Birch, A., Finch, A., Hayashi, H., Konstas, I., Luong, T., Neubig, G., Oda, Y., Sudoh, K. (eds.) Proceedings of the 3rd Workshop on Neural Generation and Translation. pp. 157--167. Association for Computational Linguistics, Hong Kong (Nov 2019). \doi{10.18653/v1/D19-5616}, \url{https://aclanthology.org/D19-5616}

\bibitem{shang2023post}
Shang, Y., Yuan, Z., Xie, B., Wu, B., Yan, Y.: Post-training quantization on diffusion models. In: Proceedings of the IEEE/CVF Conference on Computer Vision and Pattern Recognition. pp. 1972--1981 (2023)

\bibitem{so2023temporal}
So, J., Lee, J., Ahn, D., Kim, H., Park, E.: Temporal dynamic quantization for diffusion models. In: Thirty-seventh Conference on Neural Information Processing Systems (2023), \url{https://openreview.net/forum?id=D1sECc9fiG}

\bibitem{sohldickstein2015deep}
Sohl-Dickstein, J., Weiss, E.A., Maheswaranathan, N., Ganguli, S.: Deep unsupervised learning using nonequilibrium thermodynamics (2015)

\bibitem{song2020denoising}
Song, J., Meng, C., Ermon, S.: Denoising diffusion implicit models. International Conference on Learning Representations  (2020), \url{dblp.org/rec/journals/corr/abs-2010-02502}

\bibitem{wang2024quest}
Wang, H., Shang, Y., Yuan, Z., Wu, J., Yan, Y.: Quest: Low-bit diffusion model quantization via efficient selective finetuning (2024)

\bibitem{xie2022measurement}
Xie, Y., Li, Q.: Measurement-conditioned denoising diffusion probabilistic model for under-sampled medical image reconstruction. In: International Conference on Medical Image Computing and Computer-Assisted Intervention. pp. 655--664. Springer (2022)

\bibitem{yang2023diffusion}
Yang, L., Zhang, Z., Song, Y., Hong, S., Xu, R., Zhao, Y., Zhang, W., Cui, B., Yang, M.H.: Diffusion models: A comprehensive survey of methods and applications. ACM Computing Surveys  \textbf{56}(4),  1--39 (2023)

\bibitem{yu15lsun}
Yu, F., Zhang, Y., Song, S., Seff, A., Xiao, J.: Lsun: Construction of a large-scale image dataset using deep learning with humans in the loop. arXiv preprint arXiv:1506.03365  (2015)

\bibitem{zhang2023adding}
Zhang, L., Rao, A., Agrawala, M.: Adding conditional control to text-to-image diffusion models. In: Proceedings of the IEEE/CVF International Conference on Computer Vision. pp. 3836--3847 (2023)

\end{thebibliography}

\clearpage
\appendix
{
\centering
\Large
\textbf{Appendix}\\[1em] 
}

\section{Implementation Details}
\label{sec:implementation_details}

For CIFAR-10 image generation~\cite{krizhevsky2009learning}, we employ the 100-step DDIM approach~\cite{song2020denoising}. For the LSUN-Bedroom and LSUN-Church datasets~\cite{yu15lsun}, we implement 200 steps with LDM-4 and 500 steps with LDM-8~\cite{rombach2022high}, respectively. In conditional image generation, we use the official pre-trained Stable Diffusion version 1.4~\cite{rombach2022high}, generating images with both 50-step PLMS and DDIM samplers. We adopt methods from~\cite{li2023q,li2021brecq,nagel2020up,esser2020learned} for model quantization and calibration, and use code from~\cite{li2023q} to quantize model. 

To calculate the reconstruction coefficients and input bias, we first run the full-precision model to generate a batch of samples, capturing the input and noise estimations at each timestep. This is followed by running the quantized model to determine these coefficients. Batch sizes are tailored to each task: 64 for CIFAR-10, 128 for LSUN experiments, and 256 for text-guided image generation with Stable Diffusion v1.4. In general, larger sample size may lead to better results. we leave this for future investigation. 

We evaluate the FID score~\cite{heusel2017gans} using the official PyTorch implementation. For the IS score~\cite{salimans2016improved} evaluation on CIFAR-10, we utilize code from~\cite{dhariwal2021diffusion}. For high-resolution datasets like LSUN-Bedroom and LSUN-Church, we efficiently assess the results using pre-computed statistics over the entire dataset, as provided by~\cite{dhariwal2021diffusion}. For comparative experiments, we rerun the official scripts from \cite{shang2023post,li2023q,he2023ptqd}.

\section{Comparison of Input Bias Correction and Noise Estimation Correction}
\label{sec:comparison_ibc}
\begin{figure}[h]
  \centering
  \begin{subfigure}{\textwidth}
    \includegraphics[width=\linewidth]{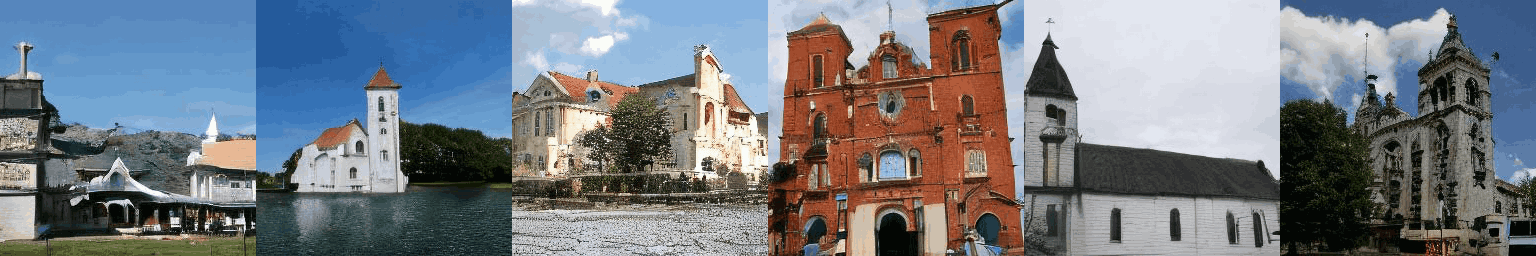}
    \caption{FP32}
  \end{subfigure}
  \hfill
  \begin{subfigure}{\textwidth}
    \includegraphics[width=\linewidth]{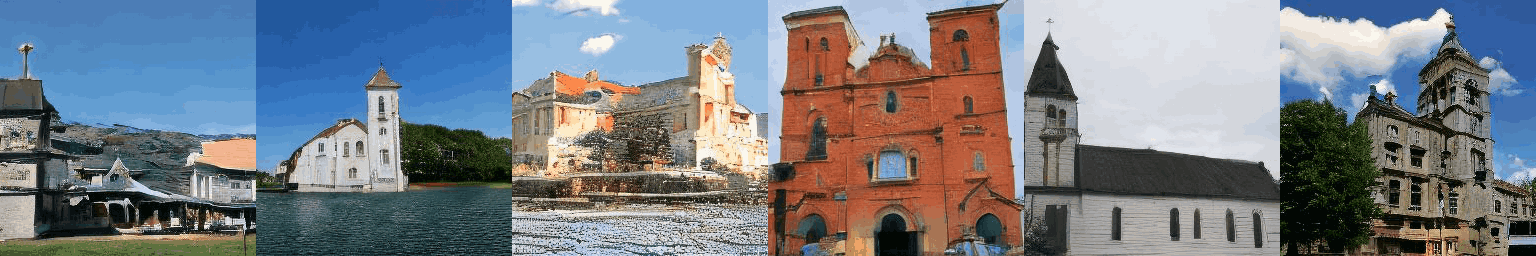}
    \caption{Input Bias Correction}
  \end{subfigure}
  \hfill
  \begin{subfigure}{\textwidth}
    \includegraphics[width=\linewidth]{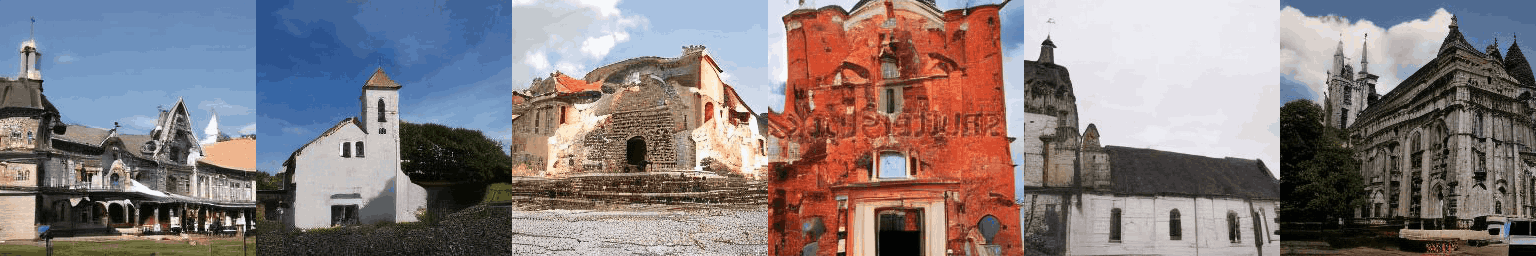}
    \caption{Noise Estimation Bias Correction}
  \end{subfigure}
  \caption{Comparison between different correction strategies in 256 $\times$ 256 unconditional generation on LSUN-Church with W3A8 500 steps LDM-8}
  \label{fig:correction_strategies}
\end{figure}

In this section, we perform a comparative analysis between Input Bias Correction (IBC), as introduced in Section \ref{Sec:ibc}, and the noise estimation bias correction approach inspired by~\cite{nagel2019data}. While the former method simultaneously corrects both the estimated noise, $\boldsymbol{\hat{\epsilon}}_t$, and the corrupted input, $\mathbf{\hat{x}_t}$, the latter focuses exclusively on correcting the corrupted noise estimation, $\boldsymbol{\hat{\epsilon}}_t$. The visualization results, presented in Figure \ref{fig:correction_strategies}, clearly demonstrate that the noise estimation correction strategy is less effective at preserving original content, often resulting in the loss of important objects and causing structural distortions in the generated images. Conversely, the IBC strategy, as implemented in TAC-Diffusion, produces images that are more closely aligned with those generated by the full-precision model. This efficacy can be attributed to IBC's ability to adjust the deviated model input back onto the correct path, consistent with the analysis of exposure bias discussed in Section \ref{Sec:ibc}.

\section{Extending TAC-Diffusion to DPM-Solver++}
\label{sec:dpm-solver}

In this section, we extend TAC-Diffusion to an advanced high-order solver,~\eg DPM-Solver++~\cite{lu2023dpmsolver}. The procedure for this integration is summarized in~\cref{alg:tac_dpm_solver}, where we exclude the pre-calculation process for simplicity.

To align with the notation used in~\cite{lu2023dpmsolver}, we define $\lambda_t = \log\left(\frac{\alpha_t}{\sigma_t}\right)$ within~\cref{alg:tac_dpm_solver}. Here, $\sigma_t$ represents the square root of the predefined forward variance schedule, and $\alpha_t = \sqrt{1-\sigma_t^2}$. During the sampling phase, we iterate $i$ backward from $M$ to $1$, identifying intermediate timesteps $s_i$ that fall between $t_{i-1}$ and $t_i$, thus ensuring a sequence $t_0 > s_1 > t_1 > \cdots > t_{M-1} > s_M > t_M$. To evaluate the performance of integrating TAC-Diffusion with DPM-Solver++, we conduct experiment on CIFAR-10, comparing our results with those of Q-Diffusion~\cite{li2023q}. The quantitative results are detailed in~\cref{tab:dpm_cifar}.

\begin{algorithm}[h]
\caption{Timestep-Aware Correction with DPM-Solver++(2S) Sampler}
\begin{algorithmic}[1]
\State \textbf{Input:} Quantized noise estimation model $\hat{\boldsymbol{\epsilon}}_{\theta}$, data prediction model $\boldsymbol{x}_{\theta}$, reconstruction coefficient $\boldsymbol{K}$ and initial data $\boldsymbol{x}_T$ 
\State \textbf{Output:} Corrected Output $\boldsymbol{\tilde{x}}_0$ 
\State $\tilde{\boldsymbol{x}}_{t_0} \leftarrow \boldsymbol{x}_{T}$

\For{$i \leftarrow 1$ to $M$}
    \State $h_i \leftarrow \lambda_{t_i} - \lambda_{t_{i-1}}$
    \State $r_i \leftarrow \frac{\lambda_{s_i}-\lambda_{t_{i-1}}}{h_i}$
    \State $\tilde{\epsilon}_{\theta}\left(\tilde{\boldsymbol{x}}_{t_{i-1}},t_{i-1}\right) \leftarrow  \boldsymbol{K}_{t_{i-1}}\hat{\epsilon}_{\theta}\left(\tilde{\boldsymbol{x}}_{t_{i-1}},t_{i-1}\right)$
    \State $\boldsymbol{x}_{\theta}\left(\tilde{x}_{t_{i-1}}, t_{i-1}\right) \leftarrow \frac{1}{\alpha_{t_{i-1}}}\left(\tilde{\boldsymbol{x}}_{t_{i-1}}  - \sigma_{t_{i-1}} \tilde{\epsilon}_{\theta}\left(\tilde{\boldsymbol{x}}_{t_{i-1}},t_{i-1}\right)\right) $
    \State $\boldsymbol{u}_i \leftarrow \frac{\sigma_{s_i}}{\sigma_{t_{i-1}}} \tilde{\boldsymbol{x}}_{t_{i-1}} - \alpha_{s_i}\left(e^{-r_i h_i} - 1\right)\boldsymbol{x}_{\theta}\left(\tilde{\boldsymbol{x}}_{t_{i-1}}, t_{i-1}\right) $ 
    \State $\tilde{\boldsymbol{\epsilon}}_{\theta}\left(\boldsymbol{u}_{i}, s_i\right) \leftarrow \boldsymbol{K}_{s_i}\hat{\boldsymbol{\epsilon}}_{\theta}\left(\boldsymbol{u}_{i}, s_i\right)$
    \State $\boldsymbol{x}_{\theta}\left(\boldsymbol{u}_{i}, s_{i}\right) \leftarrow \frac{1}{\alpha_{s_i}} \left(\boldsymbol{u}_i - \sigma_{s_i} \tilde{\boldsymbol{\epsilon}}_{\theta}\left(\boldsymbol{u}_{i}, s_i\right) \right)$
    \State $\boldsymbol{D}_i \leftarrow \left(1 -\frac{1}{2r_i}\right)\boldsymbol{x}_{\theta}\left(\tilde{\boldsymbol{x}}_{t_{i-1}}, t_{i-1}\right)  + \frac{1}{2r_i}\boldsymbol{x}_{\theta}\left(\boldsymbol{u}_{i}, s_{i}\right)$
    \State $\tilde{\boldsymbol{x}}_{t_i} \leftarrow \frac{\sigma_{t_i}}{\sigma_{t_{i-1}}}\tilde{\boldsymbol{x}}_{t_{i-1}} - \alpha_{t_i}\left(e^{-h_i}-1\right)\boldsymbol{D}_i$
\EndFor
\State \textbf{return} $\tilde{\boldsymbol{x}}_{t_M}$
\end{algorithmic}
\label{alg:tac_dpm_solver}
\end{algorithm}

\begin{table}[h]
\centering
\caption{Unconditional generation results on CIFAR-10 (32 $\times$ 32), with a W3A8 diffusion model and a 50 steps DPM-Solver++}
\begin{tabular}{lcll}
\toprule
\textbf{Method} & \textbf{Bits(W/A)} &  \textbf{FID}${\downarrow}$  \\ 
\midrule
Q-Diffusion~\cite{li2023q}  & 4/32  & 5.38\\
Ours & 4/32 & 5.29 \bluetext{(-0.09)}\\
\hline
Q-Diffusion~\cite{li2023q}  & 4/8 & 10.27\\
Ours & 3/8 &   10.05 \bluetext{(-0.22)}\\
\hline
Q-Diffusion~\cite{li2023q}  & 3/8  & 38.82 \\
Ours & 3/8 & 18.70 \bluetext{(-20.12)}\\
\bottomrule
\label{tab:dpm_cifar}
\end{tabular} 
\end{table}

\section{Model Efficiency}
\label{sec:Efficiency} 
In this section, we test the efficiency of the quantized diffusion model relative to its full-precision counterpart. We employed the official PyTorch Quantization API for model quantization. Given that this API does not support quantization to precisions lower than 8-bit, we quantized both the weights and activations to 8-bit precision.~\cref{tab:latency} showcases the average inference time for a 100-step DDIM process on the CIFAR-10 dataset, conducted on an Intel Xeon Platinum 8358 CPU. Operating with a batch size of 32, the quantized diffusion model achieves a speed-up of morethan 3.9 times, while its size is diminished to about one-fourth of that of the full-precision model. Furthermore, we note that the additional computational overhead for our proposed method is minimal, resulting in a mere 0.65\% increase in inference time compared to the Q-Diffusion~\cite{li2023q} with a batch size of 32.

\begin{table}[h]
\centering
\caption{Inference speed test on CIFAR-10 (32 $\times$ 32), with pixel-space DDIM.}
\begin{tabular}{ccccccc}
\toprule
Model & Method & Batch Size & Bits (W/A) & Size (Mb) & Time(s) & Acceleration($\times$) \\ 
\midrule
\multirow{12}{*}{\parbox{2cm}{DDIM \\ (steps = 100 \\ eta = 0.0 )}} & Full-Precision & 64 & 32/32 & 143.20  & 77.95 & 1 \\
\cline{2-7}
& Q-Diffusion~\cite{li2023q}& 64 & 8/8 & 36.21 & 26.79 & 2.91 \\ 
\cline{2-7}
& Ours & 64 & 8/8 & 36.21& 26.98 & 2.89 \\ 
\cline{2-7}
& Full-Precision & 32 & 32/32 & 143.20  & 36.18 & 1 \\
\cline{2-7}
& Q-Diffusion~\cite{li2023q}& 32 & 8/8 & 36.21 &  9.17 & 3.95  \\ 
\cline{2-7}
& Ours & 32 & 8/8 & 36.21 & 9.23 & 3.92 \\ 
\cline{2-7}
& Full-Precision & 16 & 32/32 & 143.20 & 13.48 & 1 \\ 
\cline{2-7}
& Q-Diffusion~\cite{li2023q} & 16 & 8/8 & 36.21  & 5.86 & 2.30 \\ 
\cline{2-7}
& Ours & 16 & 8/8 & 36.21 & 6.03 & 2.24 \\ 
\cline{2-7}
& Full-Precision & 1 & 32/32 & 143.20 & 3.59 & 1 \\ 
\cline{2-7}
& Q-Diffusion~\cite{li2023q} & 1 & 8/8 & 36.21  & 2.69 & 1.33 \\ 
\cline{2-7}
& Ours & 1 & 8/8 & 36.21 & 2.76 & 1.30\\ 
\bottomrule
\label{tab:latency}
\end{tabular}
\end{table}

\section{Visualization of Dynamic Activation Distribution in Noise Estimation Network}
\label{sec:dynamic_activatoin}
\begin{figure}[h]
  \centering
  \begin{subfigure}{\columnwidth}
    \includegraphics[width=0.49\linewidth]{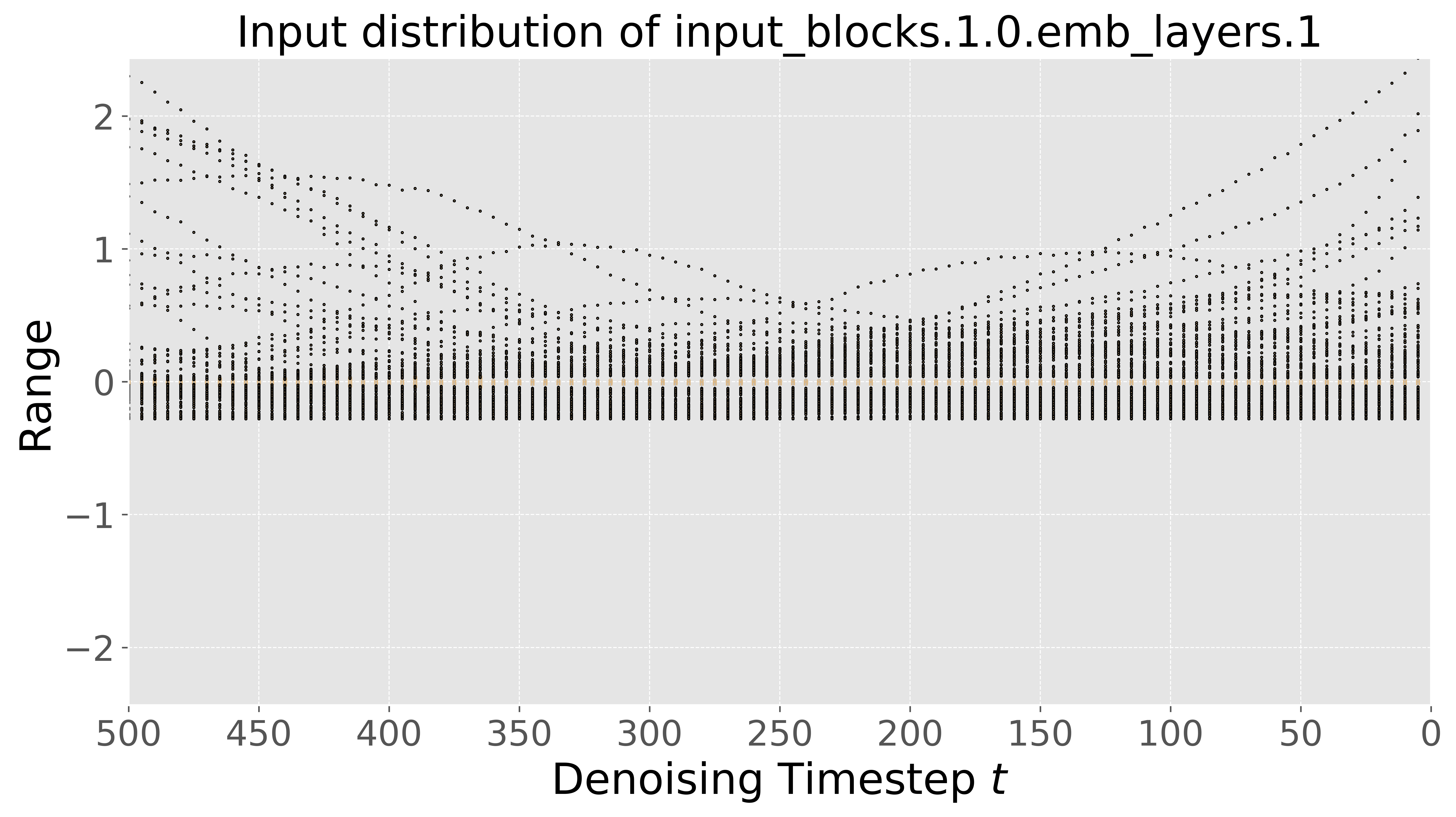}
    \includegraphics[width=0.49\linewidth]{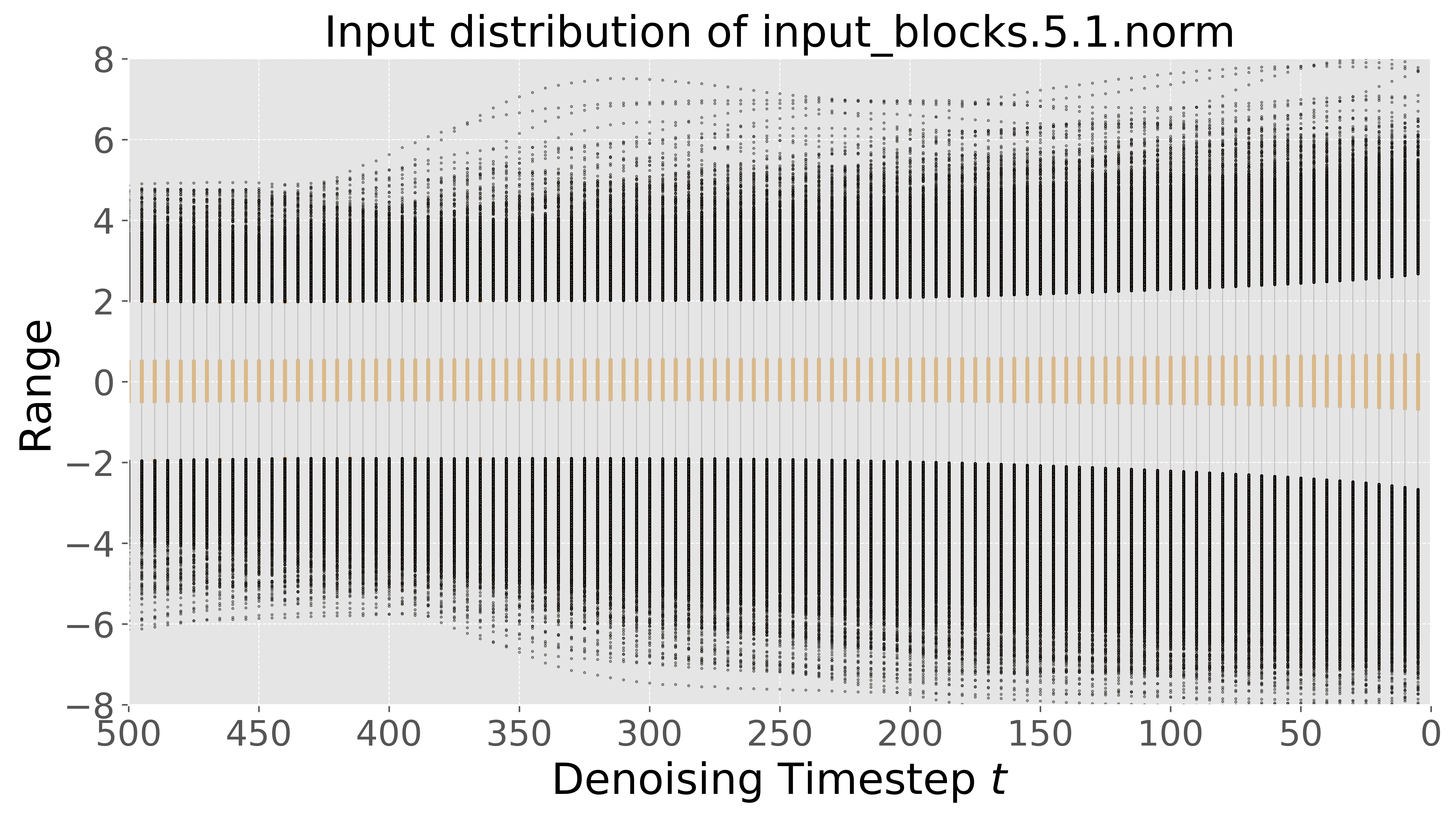}
  \end{subfigure}\\
  \begin{subfigure}{\columnwidth}
    \includegraphics[width=0.49\linewidth]{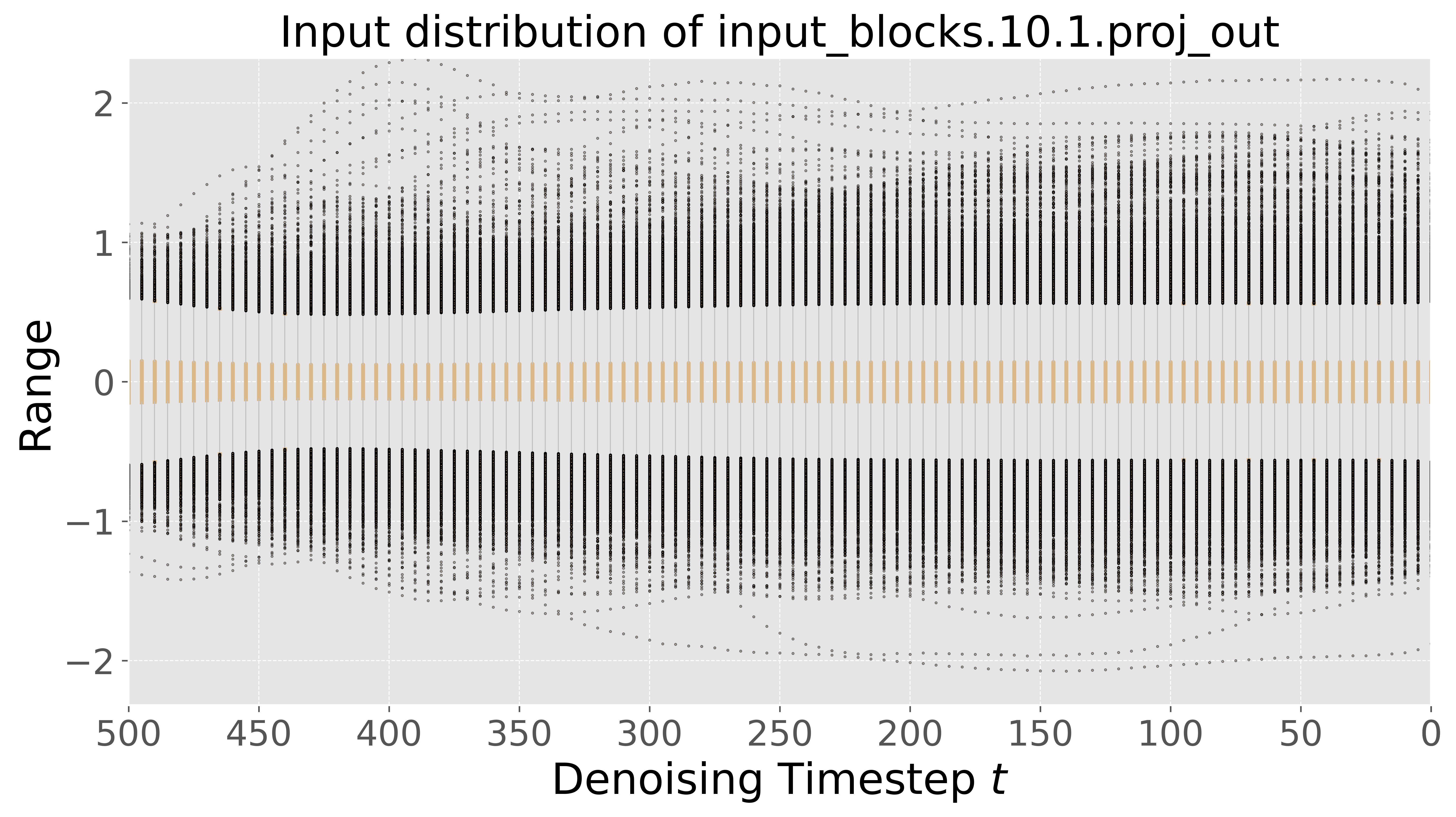}
    \includegraphics[width=0.49\linewidth]{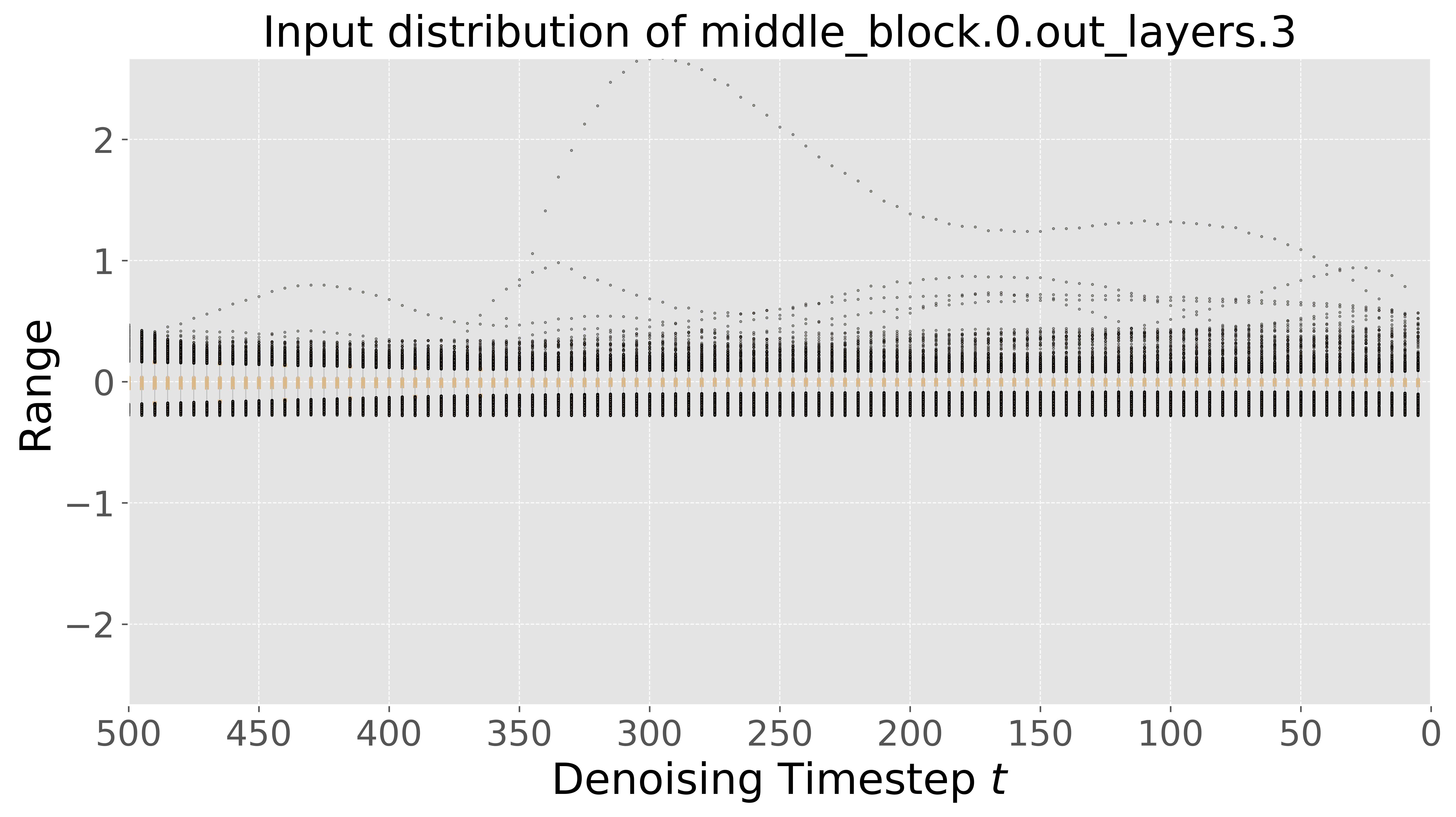}
  \end{subfigure}\\
  \begin{subfigure}{\columnwidth}
    \includegraphics[width=0.49\linewidth]{fig/sup/activation/input_blocks__10__1__proj_out.png}
    \includegraphics[width=0.49\linewidth]{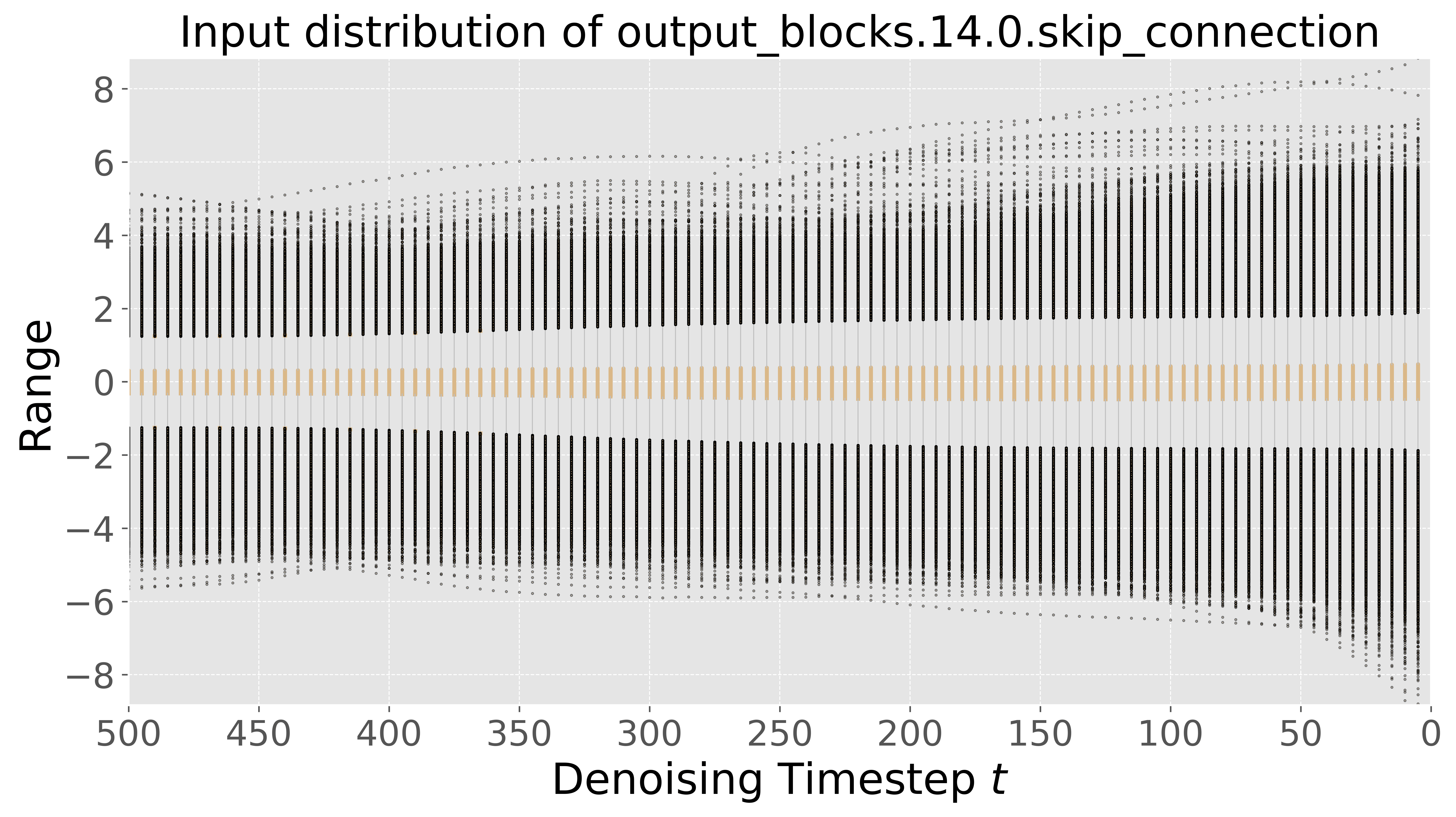}
  \end{subfigure}
  \caption{The activation distribution of multiple layers in full-precision LDM-8 on LSUN-Church. The distribution varies during the denoising process. This dynamic nature of activation is the main source of clipping error in low-precision diffusion model.}
  \label{fig:activation_distribution}
\end{figure}

We visualize the activation distribution in several layers of LDM-8 during the denoising process in~\cref{fig:activation_distribution}. We can observe that the range of activation varies greatly across timesteps in these layers. Since low-precision diffusion models maintain a fixed quantization step size, a significant portion of activation values inevitably becomes clamped during numerous timesteps. This clamping phenomenon, occurring in many timesteps, leads to substantial information loss.

\section{Ablation Study on rQSNR weight in the Reconstruction Loss Function}
\label{sec:albation_rqsnr}

\begin{figure}[h]
    \centering  
    \includegraphics[width=0.7\linewidth]{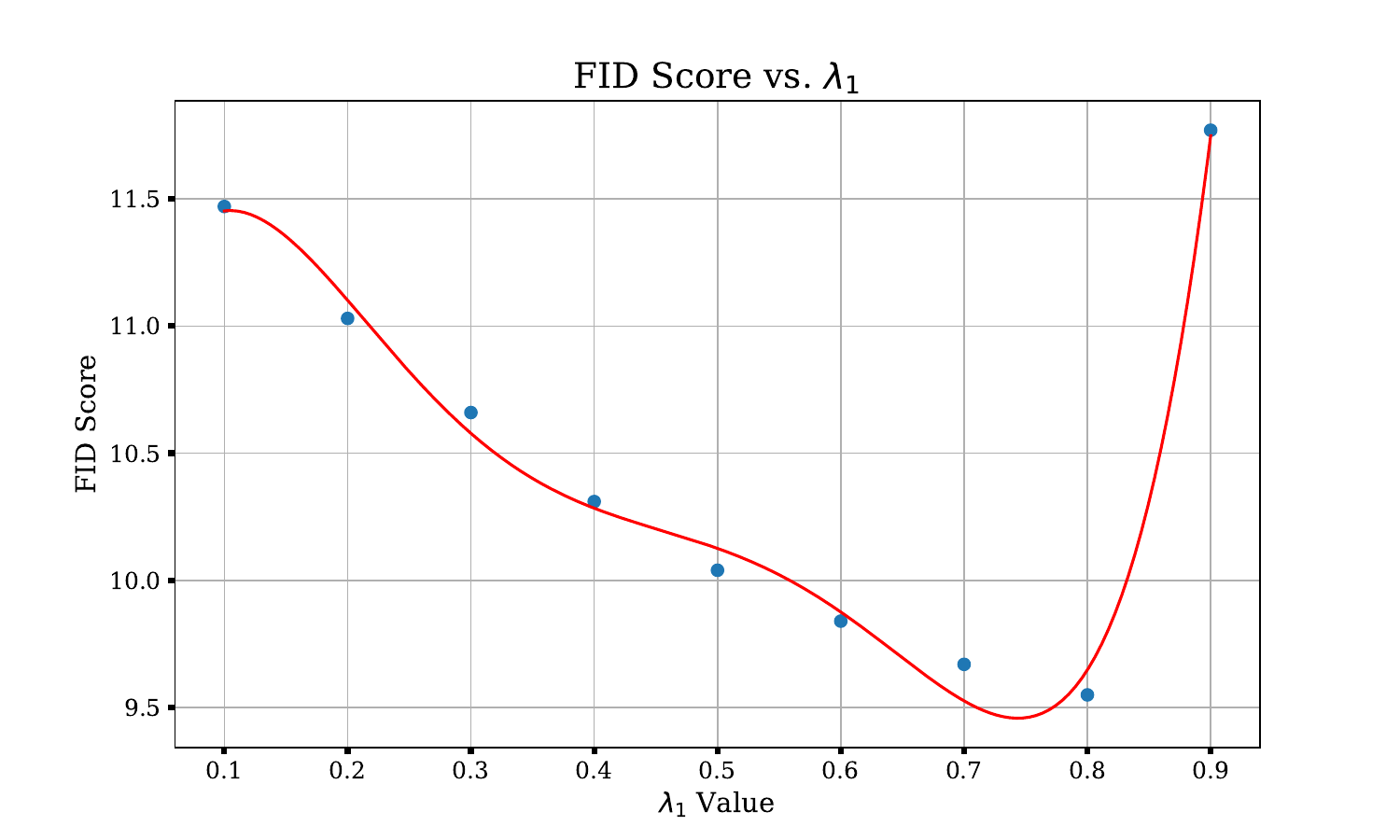}
  \caption{Ablation study on rQSNR weight}
  \label{fig:ablation_rqsnr}
\end{figure}

In this section, we conduct an ablation study on the weighting coefficient $\lambda_1$ of the rQNSR penalty in the reconstruction loss function. This study employs a W3A8 100-step DDIM on the CIFAR-10 dataset. The analysis, illustrated in~\cref{fig:ablation_rqsnr}, explores the balance between the mean square error and the relative quantization noise sensitivity by uniformly adjusting $\lambda_1$ in increments of $0.1$. This ensures that both $\lambda_1$ and $1 - \lambda_1$ remain within positive bounds. The fitted curve to the observed data points identifies an trade-off between these two  components in the reconstruction loss function. While minimizing MSE is a prevalent strategy in numerous post-training quantization methods~\cite{Bai_2023_ICCV,chen2023data}, our findings suggest that integrating a balanced consideration of both absolute and relative error can enhance reconstruction outcomes in quantized diffusion models, thereby leading to improved noise estimation fidelity.

\section{Potential Negative Impact}
The ability to create semantically coherent and visually compelling images with ease raises concerns over the potential misuse of such technology. It can be exploited to generate fake or misleading content, including deepfakes, that can have serious ramifications in areas such as politics, security, and personal reputation. While our method enhances the fidelity of generated images on low-precision devices, it also necessitates the development and enforcement of ethical guidelines and technological solutions to detect and prevent the misuse of synthetic media.

\section{Limitations}
In this study, our primary focus is on addressing the accumulation of quantization errors introduced by the dynamic nature of diffusion models. Extensive experiments conducted on diverse datasets demonstrate that, with the input correction at each timestep, low-precision diffusion models can effectively mitigate the accumulation of quantization errors, resulting in image quality comparable to that of full-precision diffusion models. However, it is important to note that our proposed method is applied exclusively to the model's input and the noise estimation, suggesting that quantization errors may still impact the model's inference at each timestep. Therefor, a more fine-grained correction strategy, such as correction within the residual block, might further improve the performance of quantized models. Moreover, we acknowledge the potential alternative approaches for mitigating quantization errors in low-precision diffusion models,~\eg adaptive step size. We leave the exploration of these approaches as future work.

\section{Qualitative Result on CIFAR10}
\label{sec:cifar_qualitative}
\begin{figure}[h]
  \centering
    \includegraphics[width=\linewidth]{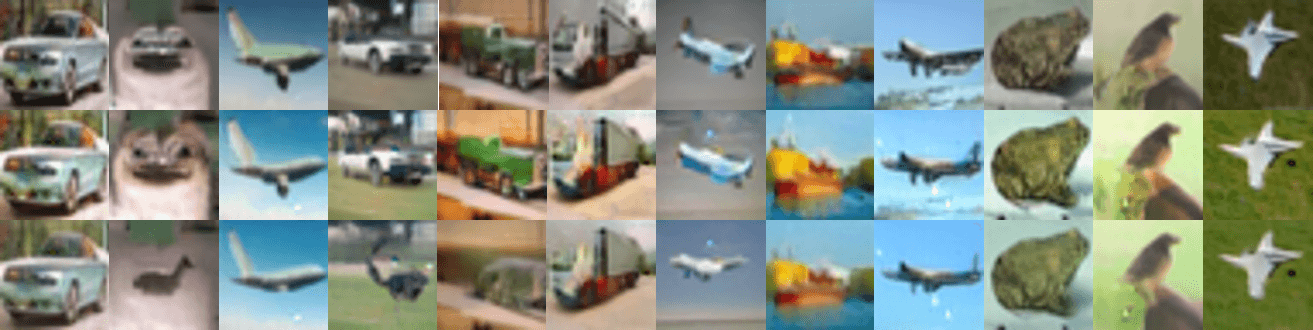}
  \caption{Unconditional image generation using the W3A8 100 steps DDIM on the CIFAR-10 dataset. The presented sequences, from top to bottom, are Full Precision model, TAC-Diffusion, and Q-Diffusion~\cite{li2023q}.}
  \label{fig:cifar}
\end{figure}

\begin{figure}[h]
  \centering
  \begin{subfigure}[b]{0.49\linewidth}
    \includegraphics[width=\linewidth]{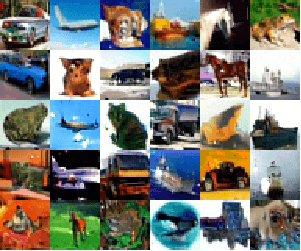}
    \caption{TAC-Diffusion}
    \label{fig:dpm38tac}
  \end{subfigure}
  \hfill
  \begin{subfigure}[b]{0.49\linewidth}
    \includegraphics[width=\linewidth]{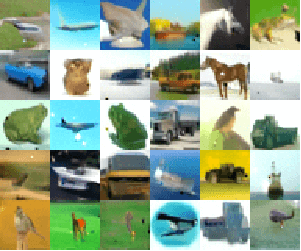}
    \caption{Q-Diffusion}
    \label{fig:dpm38qd}
  \end{subfigure}
  \caption{Unconditional image generation using the W3A8 50 steps DPM-Solver++ on the CIFAR-10 dataset.}
  \label{fig:cifar_dpm}
\end{figure}

In this section, we present the quantitative result from experiments conducted on the CIFAR-10~\cite{krizhevsky2009learning}. The generated images using the PTQ4DM~\cite{shang2023post} and Q-Diffusion~\cite{li2023q}, both implemented with W3A8 100 steps DDIM, are illustrated in~\cref{fig:cifar}. Additionally, we display results achieved with the W3A8 50 steps DPM-Solver++ in~\cref{fig:cifar_dpm}.

\section{Results on ImageNet}
\label{sec:results_imagenet}

\begin{figure}[h]
  \centering
  \captionsetup{justification = centerlast}
    \begin{subfigure}[t]{0.15\columnwidth}
        \centering
        \includegraphics[width=\textwidth]{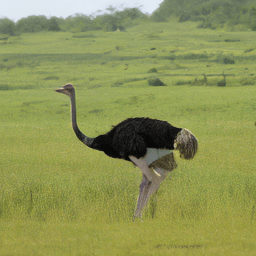}
        \caption*{\tiny FP}
    \end{subfigure}
    \begin{subfigure}[t]{0.15\columnwidth}
        \centering
        \includegraphics[width=\textwidth]{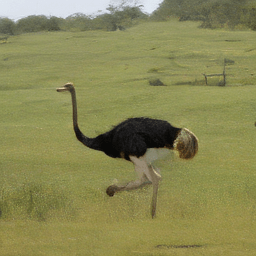}
        \caption*{\tiny Ours \\LPIPS: 0.104\\SSIM: 0.839 \\ PSNR: 25.98}
    \end{subfigure}
    \begin{subfigure}[t]{0.15\columnwidth}
        \centering
        \includegraphics[width=\textwidth]{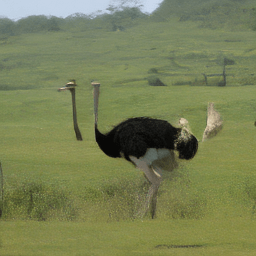}
        \caption*{\tiny TFMQ-DM~\cite{huang2023tfmq}\\LPIPS: 0.151\\SSIM: 0.829 \\ PSNR: 21.75}
    \end{subfigure}
    \begin{subfigure}[t]{0.15\columnwidth}
        \centering
        \includegraphics[width=\textwidth]{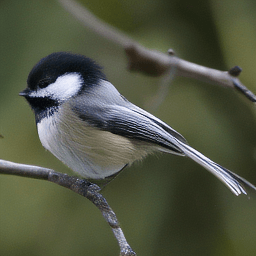}
        \caption*{\tiny FP}
    \end{subfigure}
    \begin{subfigure}[t]{0.15\columnwidth}
        \centering
        \includegraphics[width=\textwidth]{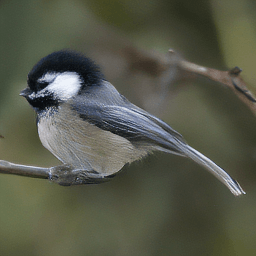}
        \caption*{\tiny Ours\\LPIPS: 0.146\\SSIM:0.848 \\ PSNR: 22.26}
    \end{subfigure}
    \begin{subfigure}[t]{0.15\columnwidth}
        \centering
        \includegraphics[width=\textwidth]{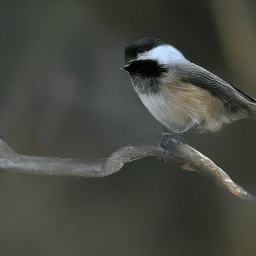}
        \caption*{\tiny TFMQ-DM~\cite{huang2023tfmq}\\LPIPS: 0.443\\SSIM: 0.649 \\ PSNR: 16.97}
    \end{subfigure}
    \label{fig:imagenet}
  \caption{Conditional generation on ImageNet (256 $\times$ 256) with W3A8 20 steps LDM-8}
\end{figure}

To further explore the performance of our method on ImageNet, we report the LPIPS, PSNR, SSIM, and FID in~\cref{tab:tfmq-dm-comparison}. Visualization results with W3A8 precision are provided in~\cref{fig:imagenet}.

\begin{table}[h]
    \caption{Conditional generation results on ImageNet (256 $\times$ 256), with 20 steps LDM-8}
    \centering
    \begin{tabular}{lccccc}
        \toprule
        \textbf{Methods} & \textbf{Bits (W/A)}& \textbf{LPIPS}${\downarrow}$ &  \textbf{PSNR}${\uparrow}$ & \textbf{SSIM}${\uparrow}$  &  \textbf{FID}${\downarrow}$  \\ 
        \midrule  
        Full-Precision & 32/32 & --  & -- & -- & 10.91\\
        \midrule
        TFMQ-DM~\cite{huang2023tfmq}  & 8/8 & 0.026  & 33.59 & 0.958 & \textbf{10.79} \\
        Ours  & 8/8 & \textbf{0.023} & \textbf{34.14} & \textbf{0.962 }&  10.82 \\
        \midrule
        TFMQ-DM~\cite{huang2023tfmq}  & 3/8 & 0.227  & 20.61 & 0.776 & 8.62\\
        Ours & 3/8 & \textbf{0.206} & \textbf{22.15} & \textbf{0.788} & \textbf{8.36}\\
        \bottomrule
    \end{tabular} 
    \label{tab:tfmq-dm-comparison}
\end{table}

While LPIPS, SSIM, and PSNR evaluate the similarity between images generated by the quantized and full-precision models, improved performance on these metrics indicates our method's ability to enhance the performance of quantized models towards that of the FP models.

\section{Quantitative Results on LSUN-Bedroom}
\label{sec:quantitative_bedroom}

\begin{figure}[h]
  \centering
  \begin{subfigure}{\textwidth}
    \includegraphics[width=\linewidth]{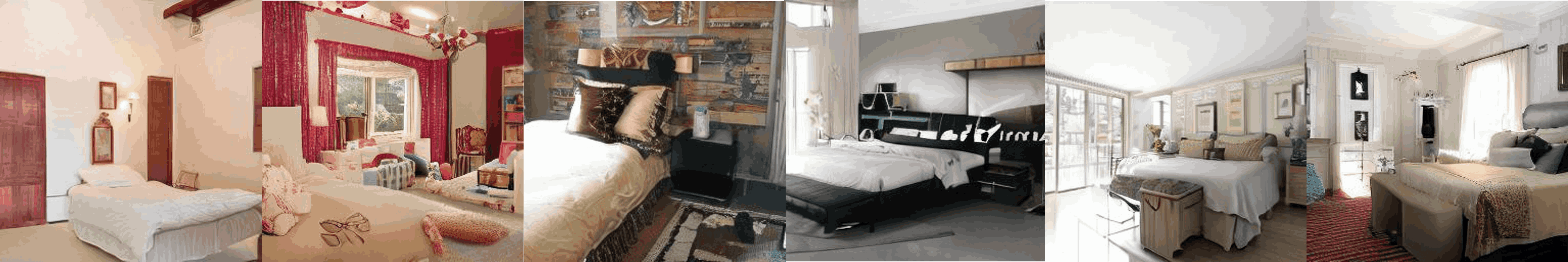}
    \includegraphics[width=\linewidth]{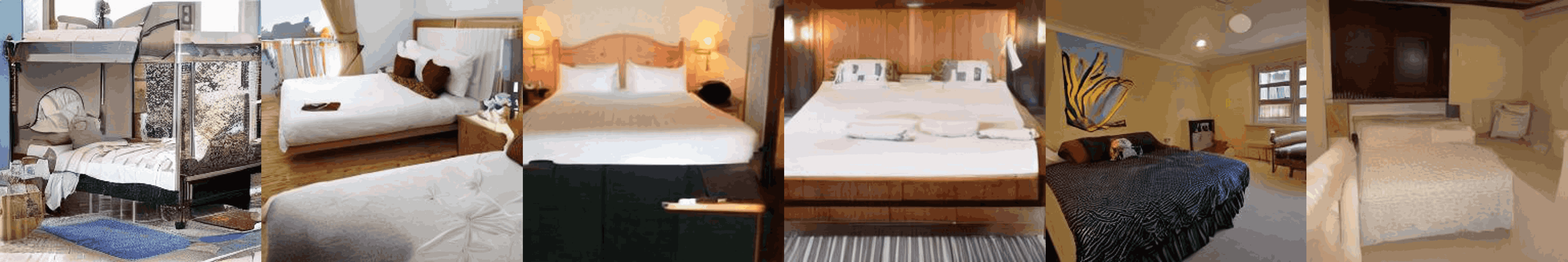}
    \caption{LDM-8 FP32}
  \end{subfigure}
  \hfill
  \begin{subfigure}{\textwidth}
    \includegraphics[width=\linewidth]{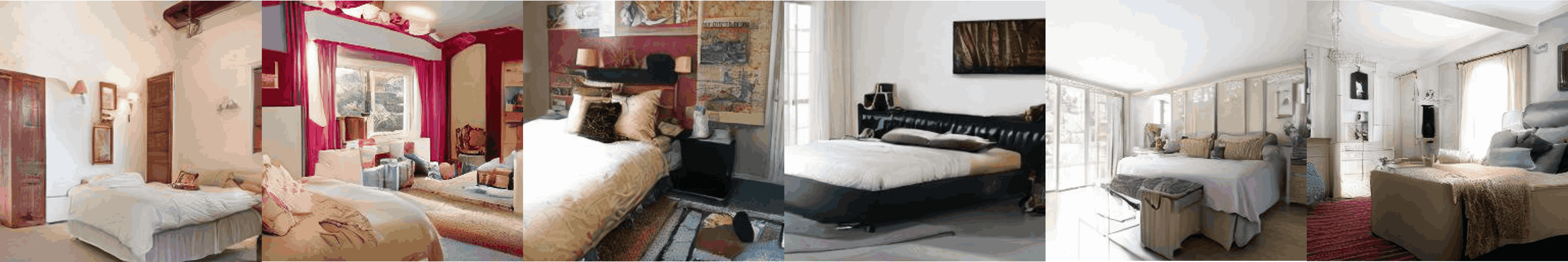}
    \includegraphics[width=\linewidth]{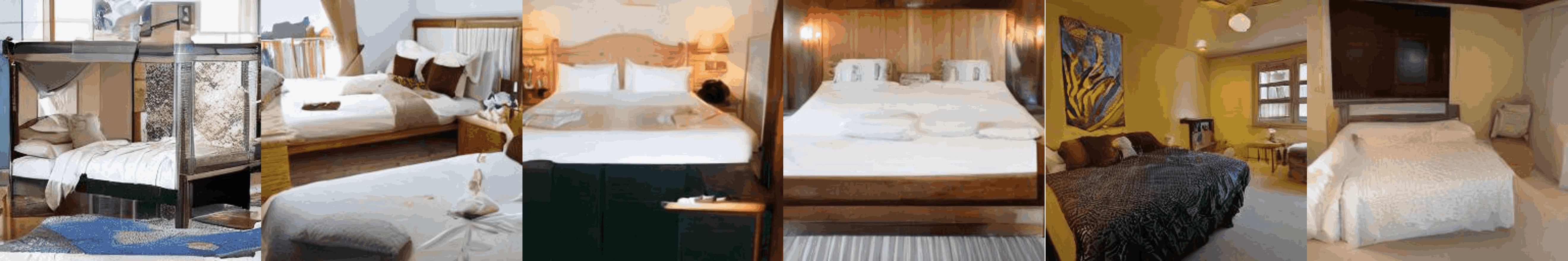}
    \caption{TAC-Diffusion W3A8}
  \end{subfigure}
  \hfill
  \begin{subfigure}{\textwidth}
    \includegraphics[width=\linewidth]{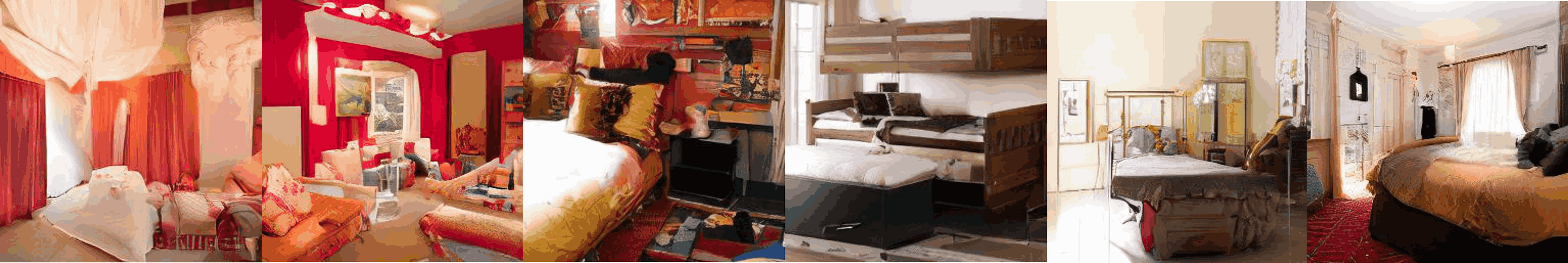}
    \includegraphics[width=\linewidth]{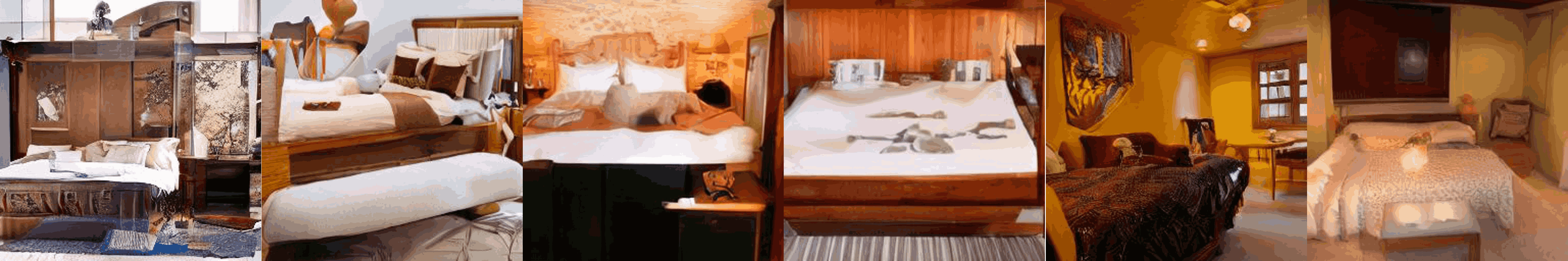}
    \caption{Q-Diffusion~\cite{li2023q} W3A8}
  \end{subfigure}
  \hfill
  \begin{subfigure}{\textwidth}
    \includegraphics[width=\linewidth]{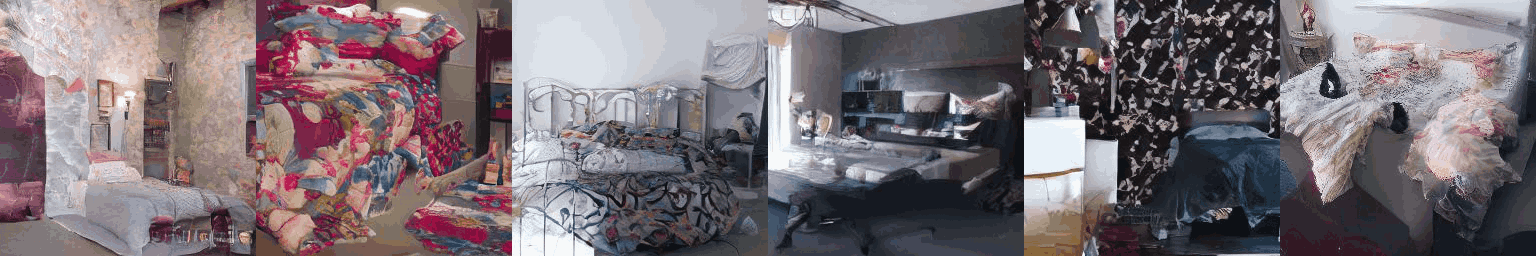}
    \includegraphics[width=\linewidth]{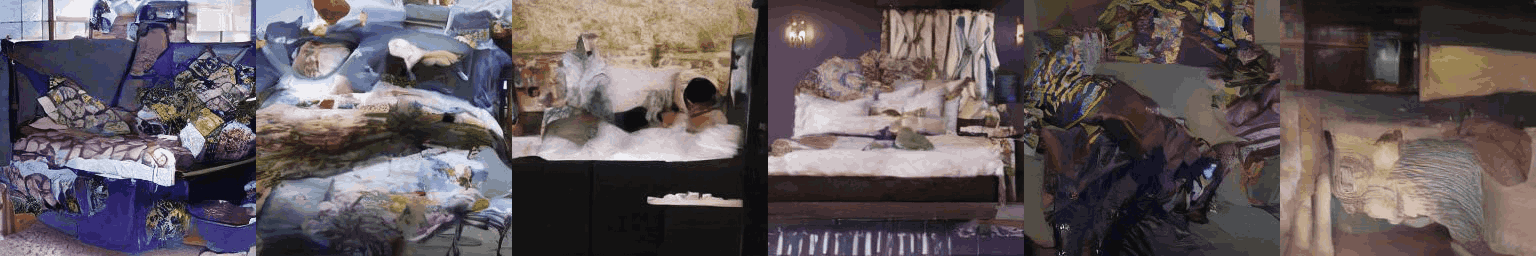}
    \caption{PTQD W3A8}
  \end{subfigure}
  \caption{256 $\times$ 256 unconditional generation on LSUN-Bedroom with W3A8 200 steps LDM-4.}
  \label{fig:sup_bed}
\end{figure}

\begin{figure}[h]
  \centering
  \begin{subfigure}{\textwidth}
    \includegraphics[width=\linewidth]{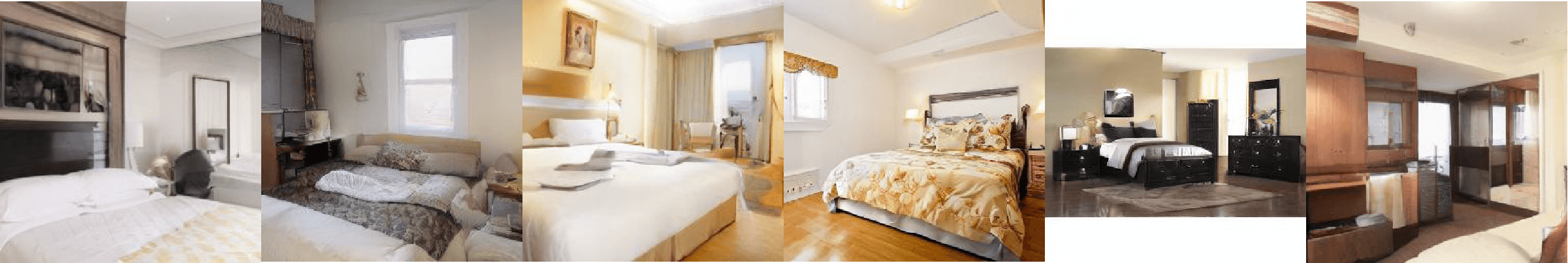}
    \caption{LDM-8 FP32}
  \end{subfigure}
  \hfill
  \begin{subfigure}{\textwidth}
    \includegraphics[width=\linewidth]{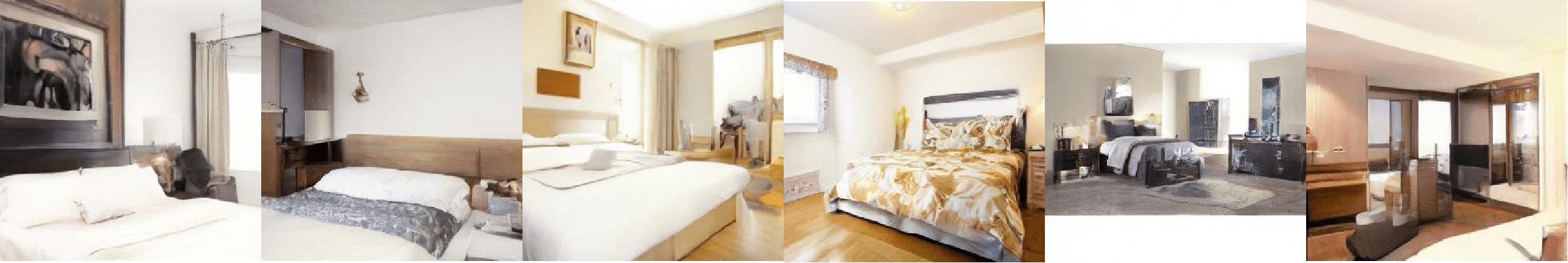}
    \caption{TAC-Diffusion W2A8}
  \end{subfigure}
  \hfill
  \begin{subfigure}{\textwidth}
    \includegraphics[width=\linewidth]{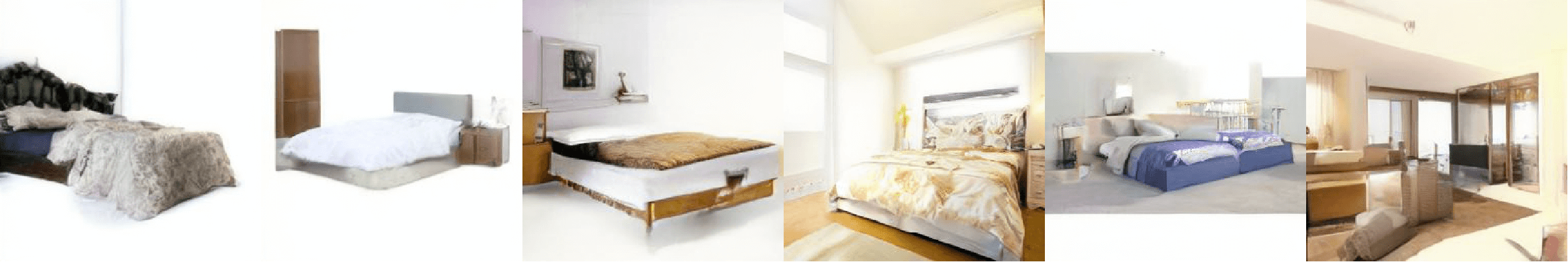}
    \caption{Q-Diffusion~\cite{li2023q} W2A8}
  \end{subfigure}
  \caption{256 $\times$ 256 unconditional generation on LSUN-Bedroom with W2A8 200 steps LDM-4.}
  \label{fig:bedroom_w2a8}
\end{figure}

In this sectiom, we provide more quantitative results with diffusion models of extremely low precision. Unconditional generation results on LSUN-Bedroom with W3A8 and W2A8 LDM-4 are visualized in~\cref{fig:bedroom_w2a8,fig:sup_bed}. A constant improvement in image quality can be observed. Notably, when the diffusion model is quantized to 2-bit, our method can still guarantee the quality of generated image. 

\section{Conditional Generation with Stabel Diffusion}
\label{sec:stable_diffusion}

\begin{figure}[h]
  \centering

  \begin{subfigure}{\textwidth}
    \includegraphics[width=\linewidth]{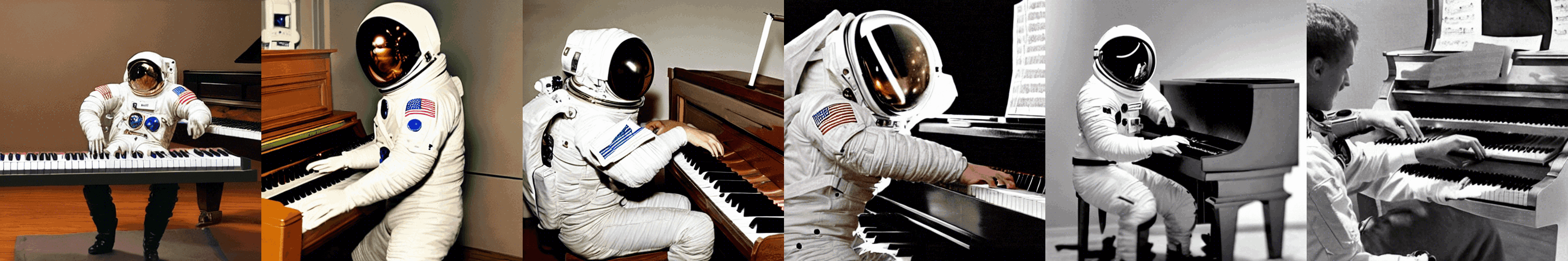}
    \caption{Stable Diffusion FP32 with 50 steps PLMS sampler}
  \end{subfigure}
  \hfill
  \begin{subfigure}{\textwidth}
    \includegraphics[width=\linewidth]{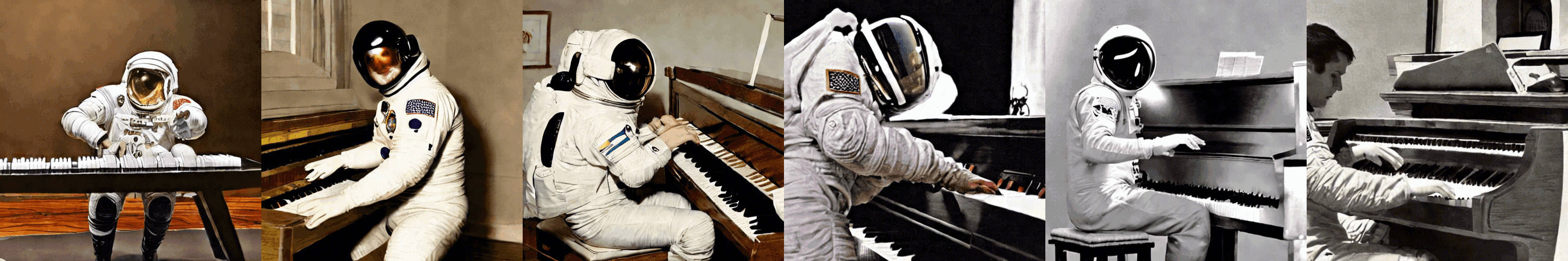}
    \caption{TAC-Diffusion W4A8 with 50 steps PLMS sampler}
  \end{subfigure}
  \hfill
  \begin{subfigure}{\textwidth}
    \includegraphics[width=\linewidth]{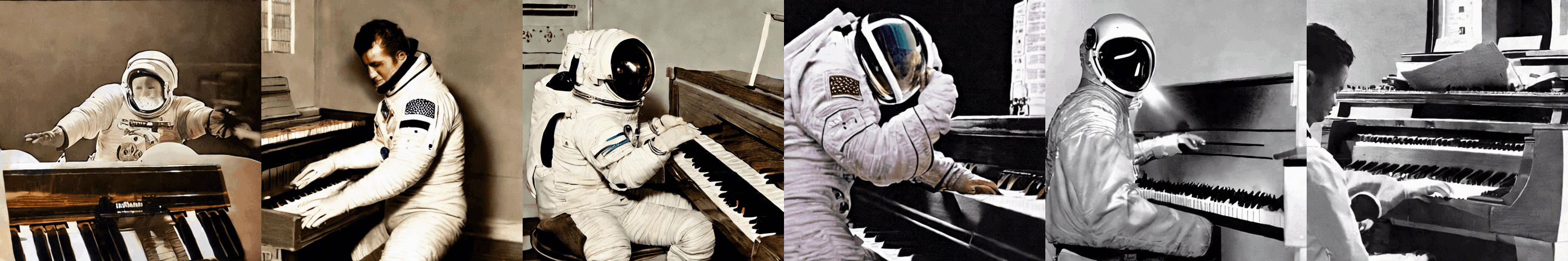}
    \caption{Q-Diffusion W4A8 with 50 steps PLMS sampler}
  \end{subfigure}
  \hfill
  \begin{subfigure}{\textwidth}
    \includegraphics[width=\linewidth]{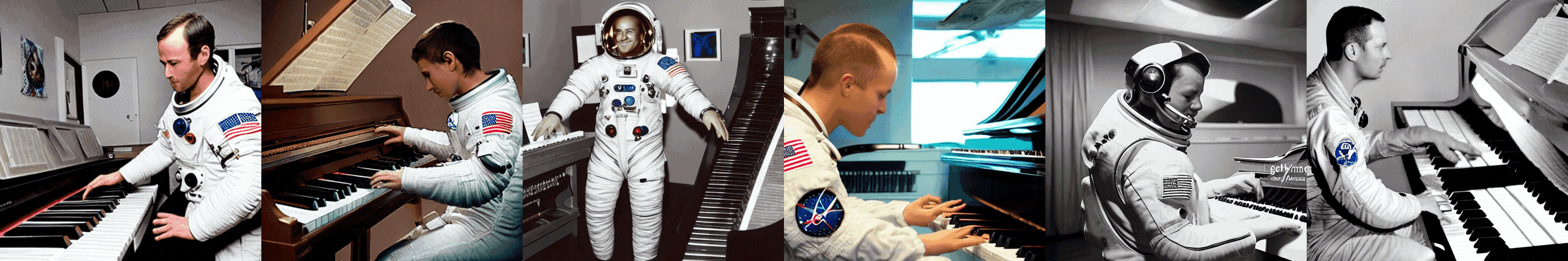}
    \caption{Stable Diffusion FP32 with 50 steps DDIM sampler}
  \end{subfigure}
  \hfill
  \begin{subfigure}{\textwidth}
    \includegraphics[width=\linewidth]{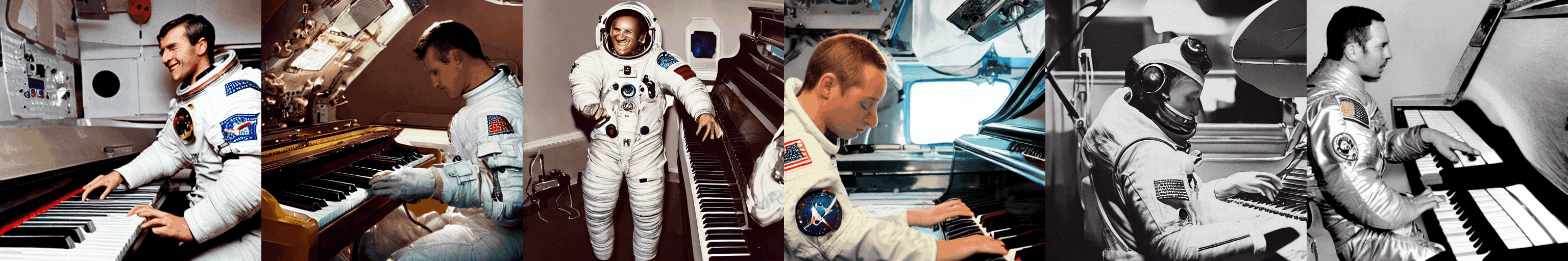}
    \caption{TAC-Diffusion W4A8 with 50 steps DDIM sampler}
  \end{subfigure}
  \hfill
  \begin{subfigure}{\textwidth}
    \includegraphics[width=\linewidth]{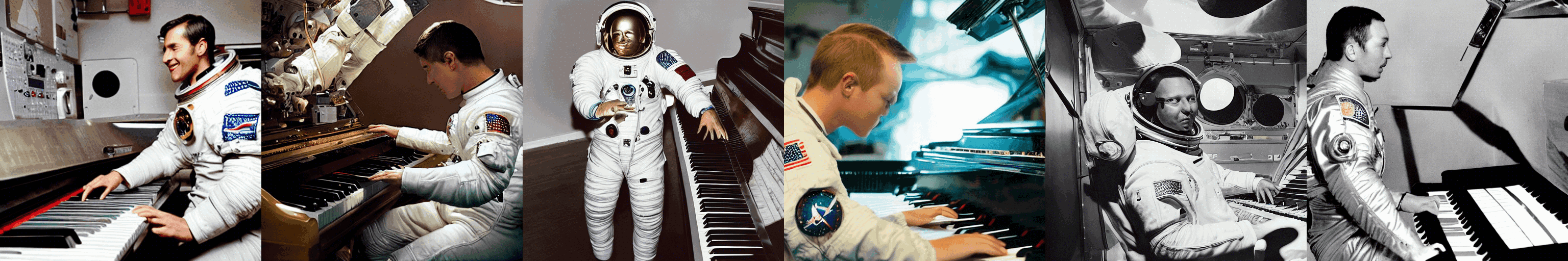}
    \caption{Q-Diffusion W4A8 with 50 steps DDIM sampler}
  \end{subfigure}

  \caption{Text-to-image generation at a resolution of 512 $\times$ 512 using Stable Diffusion, with prompt \textit{A photograph of an astronaut playing piano.}}
  \label{fig:sd_piano}
\end{figure}

\begin{figure}[h]
  \centering
  \begin{subfigure}{\textwidth}
    \includegraphics[width=\linewidth]{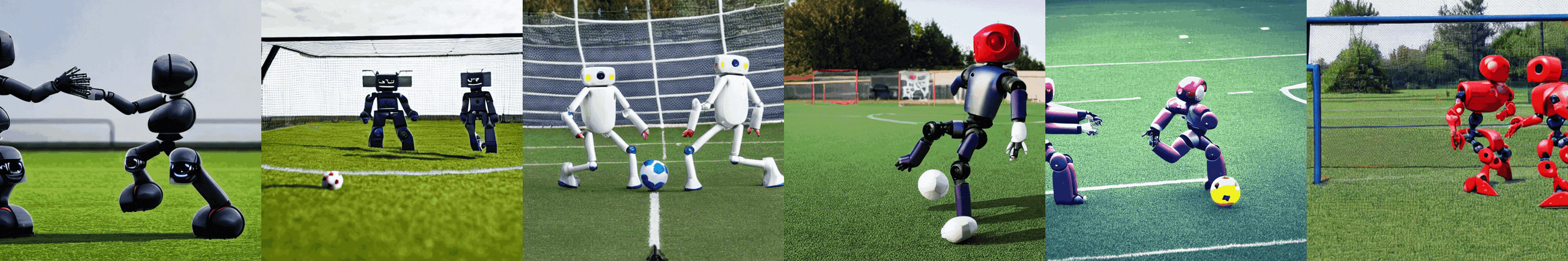}
    \caption{Stable Diffusion FP32 with 50 steps PLMS sampler}
  \end{subfigure}
  \hfill
  \begin{subfigure}{\textwidth}
    \includegraphics[width=\linewidth]{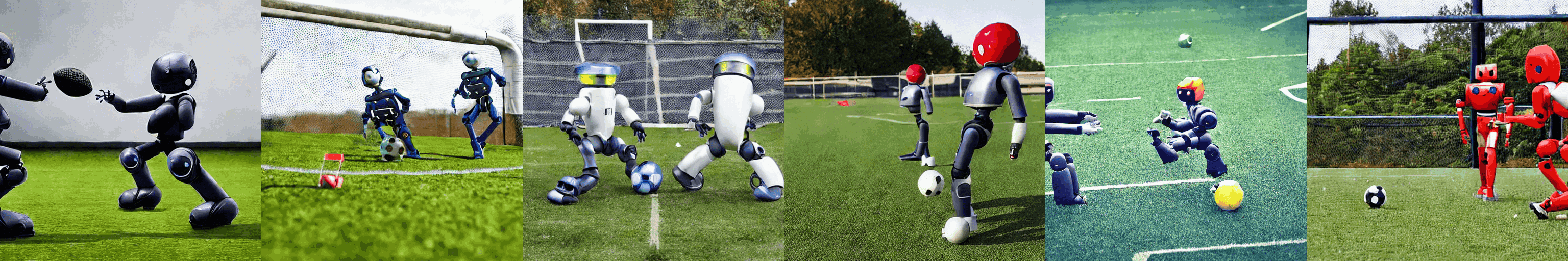}
    \caption{TAC-Diffusion W4A8 with 50 steps PLMS sampler}
  \end{subfigure}
  \hfill
  \begin{subfigure}{\textwidth}
    \includegraphics[width=\linewidth]{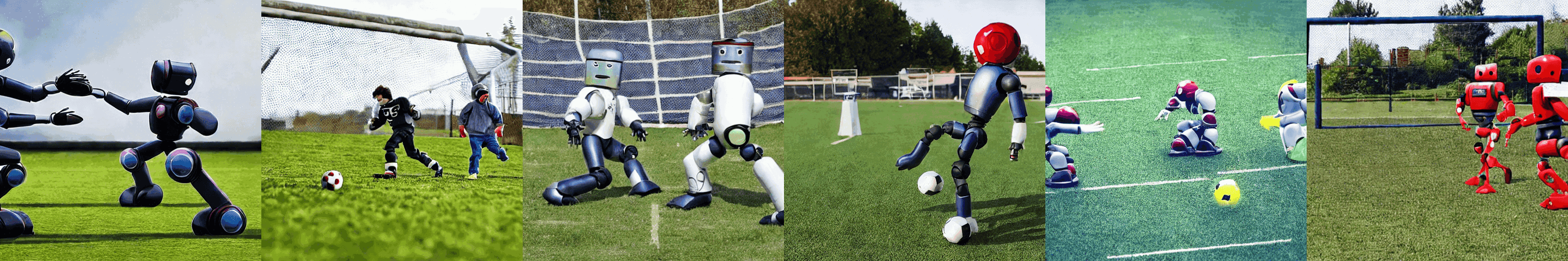}
    \caption{Q-Diffusion W4A8 with 50 steps PLMS sampler}
  \end{subfigure}
  \hfill
  \begin{subfigure}{\textwidth}
    \includegraphics[width=\linewidth]{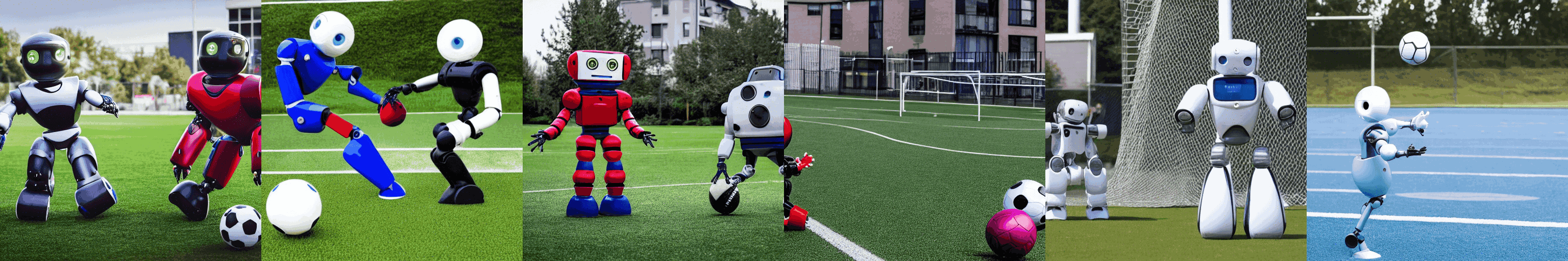}
    \caption{Stable Diffusion FP32 with 50 steps DDIM sampler}
  \end{subfigure}
  \hfill
  \begin{subfigure}{\textwidth}
    \includegraphics[width=\linewidth]{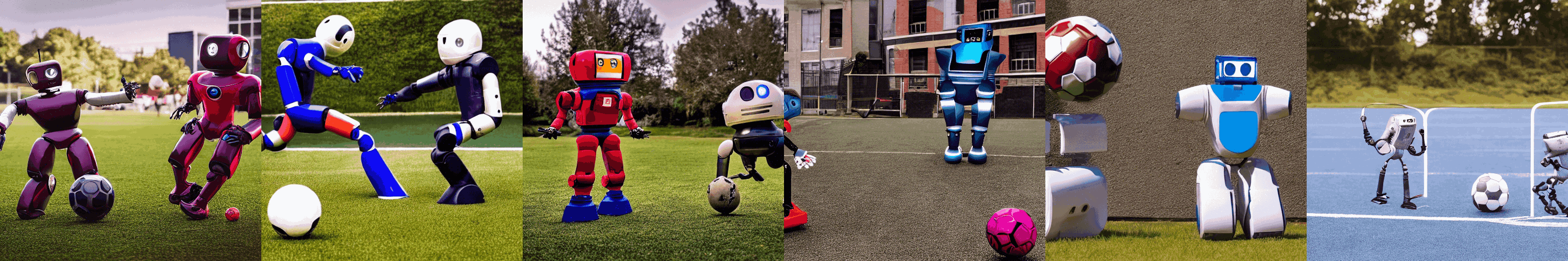}
    \caption{TAC-Diffusion W4A8 with 50 steps DDIM sampler}
  \end{subfigure}
  \hfill
  \begin{subfigure}{\textwidth}
    \includegraphics[width=\linewidth]{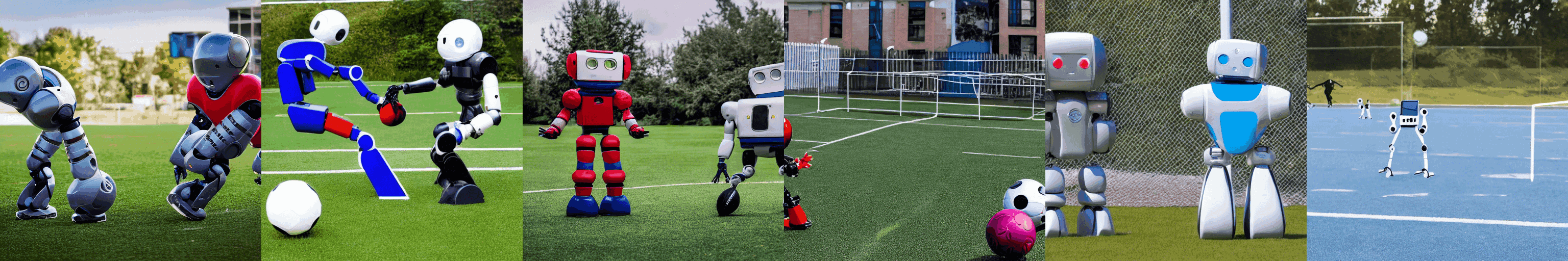}
    \caption{Q-Diffusion W4A8 with 50 steps DDIM sampler}
  \end{subfigure}
  \caption{Text-to-image generation at a resolution of 512 $\times$ 512 using Stable Diffusion, with prompt \textit{A photo of two robots playing football.}}
  \label{fig:sd_robocup}
\end{figure}

In text-guided image generation using Stable Diffusion \cite{rombach2022high}, the diffusion model provides estimates for two types of noise at each timestep: $\boldsymbol{\epsilon}_{uc}$ for unconditional noise in the input image and $\boldsymbol{\epsilon}_c$ for conditional noise, which is closely tied to the given prompt. During the collection of calibration samples for implementing our proposed method, we observed that the conditional noise associated with the input text can exhibit significant diversity when compared to unconditional noise. Consequently, a considerably larger set of prompts may be required to comprehensively capture the entire distribution of conditional noise. A practical approach is to focus solely on correcting the unconditional noise. We present images synthesized using a 50-step PLMS sampler and a 50-step DDIM sampler, as illustrated in~\cref{fig:sd_piano,fig:sd_robocup}. Compared to Q-Diffusion, our method shows remarkable improvements, especially in creating more accurate human faces and more accurately depicting the number of objects as specified in the prompt.

\end{document}